\pdfoutput=1
\documentclass[lettersize,journal]{IEEEtran}
\usepackage{amsmath,amsfonts}
\usepackage{algorithmic}
\usepackage{array}
\usepackage[caption=false,font=large,labelfont=sf,textfont=sf]{subfig}
\usepackage{textcomp}
\usepackage{stfloats}
\usepackage{url}
\usepackage{verbatim}
\usepackage{graphicx}
\usepackage{amsmath,amssymb,amsfonts}
\usepackage{bm}
\usepackage{booktabs}
\usepackage{pgffor}
\usepackage{pifont}  
\usepackage{amsfonts}
\newcommand{\cmark}{\ding{51}}
\newcommand{\xmark}{\ding{55}}
\newcommand{\FiveStar}{\(\bigstar\bigstar\bigstar\bigstar\bigstar\)}
\newcommand{\FourStar}{\(\bigstar\bigstar\bigstar\bigstar\star\)}
\newcommand{\ThreeStar}{\(\bigstar\bigstar\bigstar\star\star\)}

\newcommand{\OneStar}{\(\bigstar\star\star\star\star\)}
\usepackage{multirow}
\usepackage[hidelinks]{hyperref}
\usepackage{cite}
\usepackage{algorithm}
\usepackage{algorithmic}
\usepackage{etoolbox}

\AtBeginEnvironment{equation}{\small}
\AtBeginEnvironment{equation*}{\small}
\AtBeginEnvironment{align}{\small}
\AtBeginEnvironment{align*}{\small}
\AtBeginEnvironment{gather}{\small}
\AtBeginEnvironment{gather*}{\small}
\AtBeginEnvironment{multline}{\small}
\AtBeginEnvironment{multline*}{\small}

\def\BibTeX{{\rm B\kern-.05em{\sc i\kern-.025em b}\kern-.08em
    T\kern-.1667em\lower.7ex\hbox{E}\kern-.125emX}}
\usepackage{balance}
\title{NaviAIS: A Scenario-Level Vessel Trajectory Prediction Dataset with Vectorized Lane Priors and the NaviLane Forecasting Framework}
\author{
Yuan Gui, Hongchen Luo, Liqi Qu, Longyue Fu, and Jiao Wang%
\thanks{This work has been submitted to the IEEE for possible publication. 
Copyright may be transferred without notice, after which this version may no longer be accessible.}%
\thanks{All authors are with Northeastern University, Shenyang 110819, China. 
Corresponding author: Jiao Wang.}
}

\begin{document}
\maketitle

\begin{abstract}
Vessel trajectory prediction in complex maritime environments is essential for traffic management, collision risk warning, route planning, and autonomous navigation. Although AIS-based learning methods have achieved rapid progress, existing datasets are often released as raw message streams or irregular time series, suffering from inconsistent sampling rates, noisy observations, heterogeneous coordinate systems, and non-unified scenario construction protocols. Moreover, most public AIS resources lack structured representations of navigational lanes, waterway geometry, and navigable-region constraints, limiting reproducible and environment-aware trajectory forecasting. To address these limitations, we introduce \textit{NaviAIS}, a standardized scenario-level AIS dataset for vessel trajectory prediction. \textit{NaviAIS} organizes multi-vessel historical--future trajectories within unified temporal windows and local coordinate systems, and further provides rasterized navigable maps, vectorized lane priors, lane graphs, and structured map representations. Compared with existing vessel trajectory prediction datasets, \textit{NaviAIS} jointly supports vectorized lanes, multi-scenario coverage, vectorized maps, open accessibility, and processed trajectories. Built upon this dataset, we propose \textit{NaviLane}, a hierarchical macro-action forecasting framework for map-aware vessel trajectory prediction. \textit{NaviLane} first performs trajectory--map joint encoding to obtain a unified scene representation, then uses a discrete macro-action codebook to generate multimodal trajectory candidates in a coarse-to-refined manner. A residual refinement module improves local geometric and dynamical consistency, while a world-model-based consequence-aware evaluator further ranks candidates according to interaction risk and environmental feasibility. Experiments on \textit{NaviAIS} show that \textit{NaviLane} outperforms representative baselines in both single-modal and multimodal prediction settings, demonstrating the effectiveness of structured navigational priors, hierarchical multimodal generation, and consequence-aware candidate evaluation. 
\end{abstract}


\section{Introduction}
\IEEEPARstart{W}{ith} the continuous growth of global maritime transportation, vessel trajectory prediction has become increasingly important in maritime traffic management, collision risk warning \cite{jia2023ragan}, route planning, and autonomous navigation. Accurate forecasting of future vessel motion is essential for understanding vessel behavior and improving the safety and efficiency of intelligent maritime systems \cite{li2023ais,yang2019big}.

\begin{figure}[htbp]
    \centering
    \includegraphics[width=1.0\linewidth]{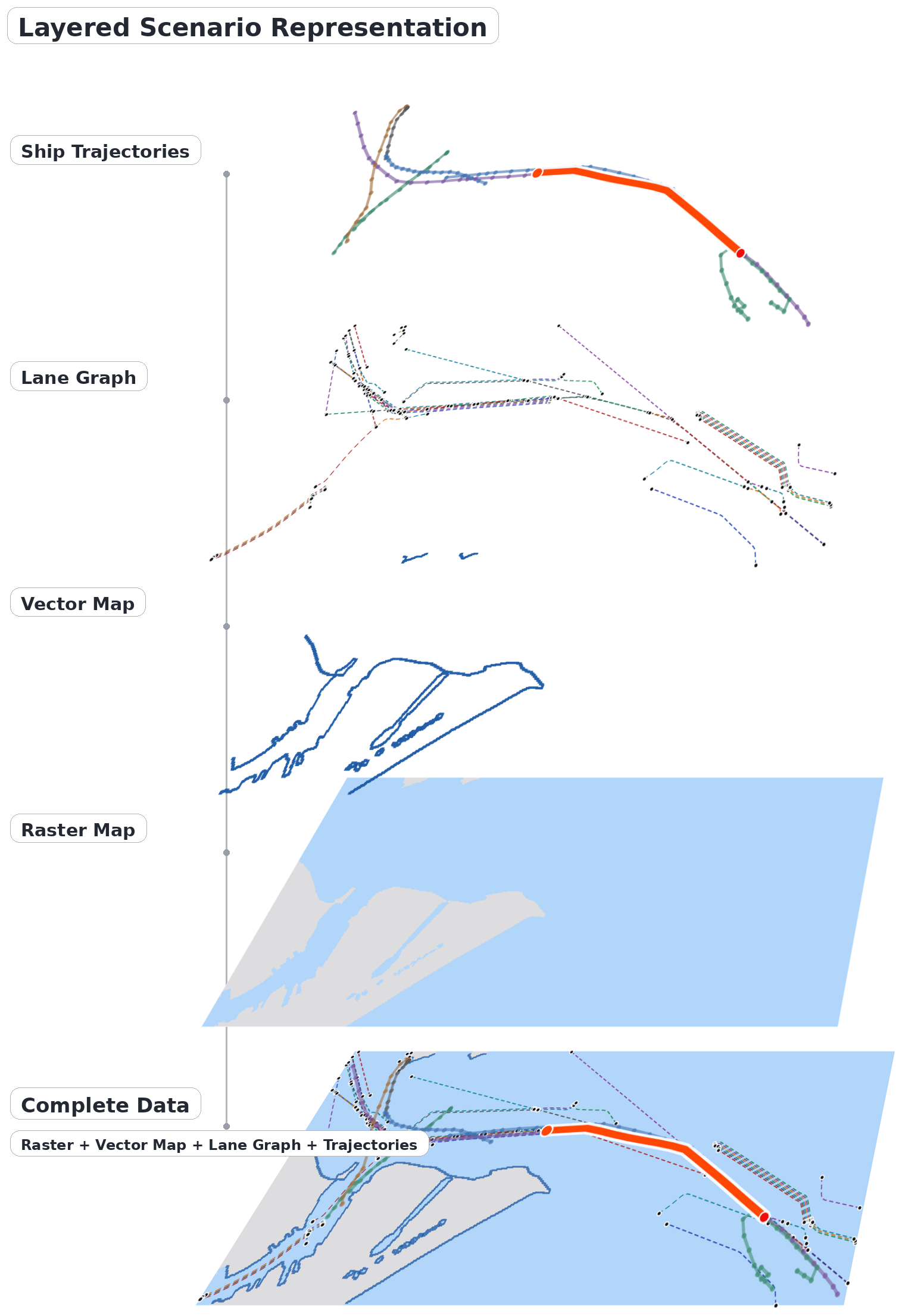}
    \vspace{-0.15cm}
    \caption{This figure illustrates the multi-layer structure of a single navigation scenario in the proposed dataset, including vessel trajectories, lane graphs, vector maps, raster maps, and the complete scene representation obtained by integrating multi-source information. By jointly representing AIS trajectories, lane vectors, water--land raster maps, and map boundaries in a unified format, the dataset captures vessel motion patterns, navigational constraints, and complex maritime geographic environments, thereby providing richer scene context for map-aware vessel trajectory prediction.}
    \label{fig:dongjitu}
    \vspace*{-0.3cm}
\end{figure}
\begin{table*}[!t]
    \centering
    \caption{Comparison of different vessel trajectory prediction datasets.}
    \label{tab:dataset_comparison}
    \renewcommand{\arraystretch}{1.2}
    \setlength{\tabcolsep}{6pt}
    \begin{tabular}{lcccccc}
    \toprule
    Dataset 
    & Scale 
    & Vectorized Lanes 
    & Multi-Scenario 
    & Vectorized Map 
    & Open Source 
    & Trajectory Processing \\
    \midrule
    USCG AIS Vessel Tracks & \FiveStar & \xmark & \cmark & \xmark & \cmark & \xmark \\
    Danish Maritime Authority Historical AIS & \FourStar & \xmark & \cmark & \xmark & \cmark & \xmark \\
    OMTAD \cite{masek2021open}& \FiveStar & \xmark & \cmark & \xmark & \cmark & \cmark \\
    North Sea ocean vessel traffic dataset \cite{meyer2023dataset} & \FiveStar & \xmark & \cmark & \xmark & \cmark & \cmark \\
    DI-MTP \cite{zhang2025unified} & \OneStar & \xmark & \xmark & \xmark & \xmark & \cmark \\
    Tptrans \cite{wang2024tptrans}& \OneStar & \xmark & \xmark & \xmark & \xmark & \cmark \\
    Ours & \ThreeStar & \cmark & \cmark & \cmark & \cmark & \cmark \\
    \bottomrule
    \end{tabular}

\end{table*}
Early studies on vessel trajectory prediction were mainly based on physical and kinematic motion models, such as constant-velocity models, constant-turn-rate models, and Nomoto-type dynamics. With the large-scale accumulation of Automatic Identification System (AIS) data \cite{svanberg2019ais,wada2021literature}, data-driven approaches have gradually become dominant, including traditional machine learning methods and, more recently, deep neural architectures such as recurrent neural networks (RNNs) \cite{yang2024harnessing,slaughter2025vessel,huang2024vessel}, Transformers \cite{xue2024g}, and graph neural networks (GNNs) \cite{kim2024higher,chen2025socialmoif,wang2024vessel,shin2024deep}. Despite the rapid progress in modeling techniques, the development of standardized data resources for vessel trajectory prediction still lags behind \cite{ettinger2021large,wilson2023argoverse}.

From the modeling perspective, many existing vessel trajectory prediction methods still face three major limitations. First, methods built primarily on historical motion sequences often lack a unified trajectory--map representation, making it difficult to explicitly incorporate structured environmental constraints such as waterway geometry, lane topology, and navigable-region boundaries. As a result, their multimodal candidates may be weakly organized at the strategy level, especially in narrow channels, curved waterways, and port areas \cite{slaughter2025vessel,huang2024vessel,xue2024g,wang2024vessel,shin2024deep}. Second, even when plausible coarse future modes can be generated, existing methods usually lack an explicit refinement mechanism to correct local geometric and dynamical errors, such as endpoint drift, curvature mismatch, and motion inconsistency. This limits the accuracy and physical plausibility of long-horizon predictions \cite{guo2023toward,zhang2025unified,wang2024tptrans,gong2025uncertainty}. Third, final candidate selection is often dominated by trajectory likelihood, intention scores, or reconstruction quality, while the interaction consequences induced by each candidate future and its compatibility with environmental constraints are not explicitly evaluated in a unified manner \cite{guo2023toward,zhang2025unified,wang2024tptrans}. These limitations suggest that high-quality vessel trajectory forecasting requires a unified trajectory--map representation with structured multimodal generation, an explicit local refinement mechanism, and a consequence-aware candidate evaluation strategy.

To address these limitations, we further propose \textit{NaviLane}, a hierarchical macro-action forecasting framework with joint trajectory--map modeling. The framework first constructs a unified scene representation from vessel motion history, static map context, and scene interaction cues. A discrete macro-action codebook is then introduced to capture high-level navigational intent, allowing the model to first select plausible strategy prototypes and then decode continuous future trajectories conditioned on the selected macro-actions. To further improve prediction quality, a residual refinement module is used to correct local geometric and dynamical errors under scene and macro-action conditions. Finally, we introduce an interaction world model together with a Counterfactual Risk (CFR) assessment mechanism to evaluate the consequences induced by different candidate futures. Specifically, the world model predicts the future states of surrounding vessels, while CFR scores each candidate trajectory according to its anticipated interaction risk and environmental feasibility, providing a principled mechanism for candidate selection beyond pure imitation \cite{guo2025vessel,gong2025uncertainty}.

Most existing AIS datasets are published in the form of raw sensor streams or irregular time series \cite{last2014comprehensive,guo2023toward}. Such data often exhibit inconsistent reporting frequencies across vessels and regions, making trajectory extraction and temporal alignment highly nontrivial. Moreover, AIS data typically contain noise, outliers, missing observations, and localization errors, which require substantial preprocessing before being used for learning-based prediction. Existing studies also differ significantly in trajectory segmentation rules, temporal horizons, coordinate normalization schemes, and scenario construction strategies. As a result, experimental settings are often inconsistent across works, making fair comparison and reproducibility difficult. Another important limitation is that existing AIS datasets usually do not provide explicit structured representations of the navigational environment \cite{le2018can}, such as waterway boundaries, lane centerlines, or navigable-area constraints, which are crucial for modeling vessel behavior in complex waterways.

To address these issues, we construct and release \textit{NaviAIS}, a standardized scenario-level AIS dataset for vessel trajectory prediction, as illustrated in Fig.~\ref{fig:dongjitu}. \textit{NaviAIS} organizes each sample as a scene containing multi-vessel trajectories within a unified temporal window and a standardized local coordinate system. In addition to trajectory observations, \textit{NaviAIS} provides vectorized lane priors, including water/land polygons, rasterized navigable masks, and lane centerlines, thereby enabling explicit modeling of environmental geometry and navigational constraints. The dataset covers multiple representative environments, including open-sea routes, inland rivers, and lake regions, and supports multi-agent interaction modeling as well as environment-aware prediction. As summarized in Table~\ref{tab:dataset_comparison}, compared with existing vessel trajectory prediction datasets, \textit{NaviAIS} provides a more complete scenario-level representation by jointly supporting vectorized lanes, multi-scenario coverage, vectorized maps, open accessibility, and processed trajectories \cite{guo2023toward}.

The main contributions of this work are summarized as follows:
\begin{itemize}
    \item We construct and release \textit{NaviAIS}, a standardized scenario-level AIS dataset for vessel trajectory prediction with vectorized lane priors and diverse navigational environments.
    \item We propose a hierarchical forecasting framework, \textit{NaviLane}, which integrates trajectory--map joint encoding, macro-action selection, conditional decoding, residual refinement, and world-model-based evaluation.
    \item We introduce a structured prediction pipeline that explicitly incorporates navigational lane priors and consequence-aware candidate evaluation, improving multimodal coverage, environmental consistency, and dynamical plausibility.
\end{itemize}

\section{Related Work}

Vessel trajectory prediction has evolved from classical kinematic models to AIS-driven learning-based approaches. Recent studies have explored recurrent architectures for sequential motion modeling, attention-based networks for multi-attribute feature fusion, Transformer-style frameworks for long-horizon dependency modeling, graph-based methods for interaction-aware prediction in crowded maritime environments, and multimodal intention-aware predictors \cite{slaughter2025vessel,huang2024vessel,xue2024g,wang2024vessel,wang2024tptrans,zhang2025unified}. Meanwhile, game-theoretic ideas have been increasingly introduced into trajectory prediction and interactive planning, since they explicitly model the mutual dependence among multiple agents. Representative works include hierarchical game-theoretic planning, Transformer-based interactive prediction and planning with hierarchical game reasoning, and Nash-equilibrium-inspired multimodal trajectory prediction \cite{fisac2019hierarchical,huang2023gameformer,lidard2023nashformer}. Although these studies have demonstrated the value of structured interaction reasoning, they are mainly developed for road traffic scenarios. In contrast, our work focuses on AIS-based maritime environments and combines structured lane-aware representation, hierarchical multimodal generation, local refinement, and consequence-aware candidate evaluation within a unified vessel trajectory prediction framework.
\section{Method}

\begin{figure*}[htbp]
    \centering
    \includegraphics[width=1.0\textwidth]{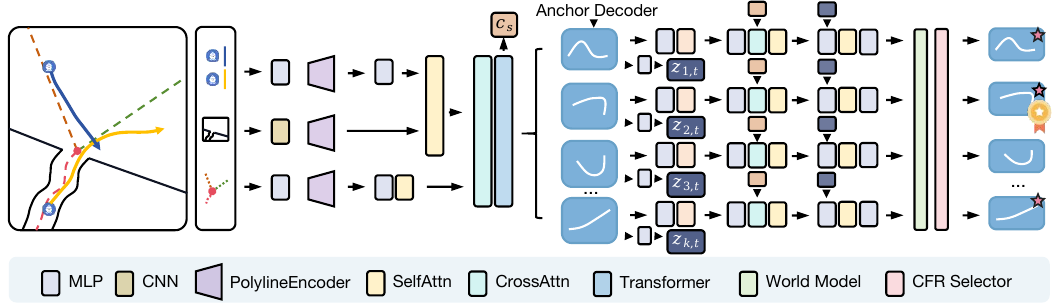}
    \caption{Framework overview: The proposed NaviLane first encodes historical vessel trajectories, vectorized lane polylines, and rasterized map priors, and fuses them with a scene interaction Transformer. Based on the fused scene context, a world model predicts surrounding-agent futures, while a macro-action anchor decoder generates multi-modal trajectory candidates in a coarse-to-refined manner. Finally, safety-aware energy, map constraints, and selector/reranker modules are jointly used to score and rank all candidates, producing the final top-$K_{\mathrm{out}}$ trajectories and the best top-1 prediction.}
    \label{fig:kuangjiatu}
\end{figure*}
This section presents a hierarchical multimodal vessel trajectory prediction framework, termed \textbf{NaviLane}. Given historical trajectories and static environmental information as inputs, the proposed framework explicitly models multimodal navigational intentions, trajectory dynamics, and environmental interactions to generate high-quality future predictions. Specifically, multi-source inputs are first encoded into a unified global scene representation. Then, a macro-action-based multimodal decoder generates multiple coarse candidate trajectories, which are further refined in parallel through conditional sequence modeling. As shown in Fig.~\ref{fig:kuangjiatu}, a hybrid evaluation strategy that combines learning-based scoring and rule-based energy modeling is finally adopted to hierarchically filter and re-rank the candidate trajectories, yielding the final prediction.

Formally, let the historical observation sequence be $\mathbf{X}=\{\mathbf{x}_t\}_{t=1}^{T_h}$, the prediction horizon be $T$, and the model generate $K_{\mathrm{macro}}$ macro-action candidate trajectories $\{\mathbf{Y}_k\}_{k=1}^{K_{\mathrm{macro}}}$, where $\mathbf{Y}_k \in \mathbb{R}^{T \times d}$. In this work, we set $K_{\mathrm{macro}}=128$ and retain $K_{\mathrm{out}}=6$ trajectories as the final multimodal outputs. The final top-1 prediction is selected from the macro-action candidates through hierarchical scoring and ranking:
\begin{equation}
\mathbf{Y}^{*} = \mathbf{Y}_{k^*}, \qquad
k^* = \arg\max_{k \in \{1,\dots,K_{\mathrm{macro}}\}} s_k,
\end{equation}
where $s_k$ denotes the comprehensive score of the $k$-th candidate trajectory. The key idea of the framework is to progressively improve candidate quality through multi-stage modeling while preserving diversity and enhancing physical plausibility and safety.

\subsection{Encoding and Trajectory Generation}

To adequately model vessel behaviors and environmental constraints, we first jointly encode historical trajectories and waterway information. The historical trajectories are mapped into agent representations $\mathbf{H}_a$ via a sequence encoder, while map information such as lane polylines and semantic attributes is encoded into map representations $\mathbf{H}_m$ through a separate map encoder:
\begin{equation}
\mathbf{H}_a = f_a(\mathbf{X}), \qquad \mathbf{H}_m = f_m(\mathbf{M}),
\end{equation}
where $\mathbf{H}_a \in \mathbb{R}^{N \times C}$ denotes the latent representations of all vessels, and $\mathbf{M}$ denotes the map input.

A Transformer-based fusion module is then employed to perform cross-modal interaction, producing a unified scene representation $\mathbf{H}_s$. A masked average pooling operation is further applied to obtain a global scene context vector $\mathbf{c}_s$:
\begin{equation}
\mathbf{H}_s = \operatorname{Fusion}(\mathbf{H}_a, \mathbf{H}_m),
\qquad
\mathbf{c}_s = \operatorname{Pool}(\mathbf{H}_s).
\end{equation}

To model multimodal futures, an anchor-driven decoding mechanism is adopted. Let $\{\mathbf{A}_k\}_{k=1}^{K_{\mathrm{macro}}}$ denote the anchor set, where each anchor corresponds to a macro-level navigation mode with a relative trajectory template $\mathbf{R}_k \in \mathbb{R}^{T \times d}$. The current state $\mathbf{x}_0$ and velocity $\mathbf{v}_0$ of the target vessel are extracted as conditional inputs and transformed into the global coordinate system through a coordinate transformation function $\mathcal{T}(\cdot)$:
\begin{equation}
\mathbf{B}_k = \mathcal{T}(\mathbf{R}_k; \mathbf{x}_0, \mathbf{v}_0),
\end{equation}
where $\mathbf{B}_k$ denotes the base trajectory of the $k$-th mode.

Conditioned on the macro-action anchors, time-dependent queries are constructed and cross-attended with the scene representation to extract contextual information and predict trajectory residuals $\Delta \mathbf{Y}_k$. The coarse trajectory of the $k$-th mode is then given by
\begin{equation}
    \Delta \mathbf{Y}_k = f_{\text{dec}}(\mathbf{A}_k, \mathbf{c}_s, \mathbf{H}_s),
    \qquad
    \mathbf{Y}_k^{(0)} = \mathbf{B}_k + \Delta \mathbf{Y}_k.
    \end{equation}
In addition, the decoder outputs a prior score for each mode,
\begin{equation}
s_k^{\text{prior}} = g(\mathbf{A}_k, \mathbf{c}_s),
\end{equation}
which reflects the prior plausibility of different navigational intentions. The $K_{\mathrm{macro}}=128$ candidate trajectories generated at this stage capture macro-level planning strategies, and the top $K_{\mathrm{out}}=6$ trajectories are retained after scoring and re-ranking.

\subsection{Trajectory Refinement}

To improve the dynamic feasibility and environmental consistency of the candidate trajectories, a refinement module is designed to process all $K_{\mathrm{macro}}$ candidates in parallel. This module applies residual corrections through a shared network structure, thereby improving trajectory quality while preserving multimodal diversity.
\begin{algorithm}[t]
    \caption{\textbf{Trajectory Refinement}}
    \label{alg:refine}
    \footnotesize
    \begin{algorithmic}[1]
    \REQUIRE Coarse trajectories $\{\mathbf{Y}_k^{(0)}\}_{k=1}^{K_{\mathrm{macro}}}$, scene representation $\mathbf{H}_s$, mode embeddings $\{\boldsymbol{m}_k\}_{k=1}^{K_{\mathrm{macro}}}$, decoder hidden states $\{\mathbf{c}_{k,t}\}$
    \ENSURE Refined trajectories $\{\mathbf{Y}_k\}_{k=1}^{K_{\mathrm{macro}}}$
    \FOR{$k = 1$ to $K_{\mathrm{macro}}$}
        \FOR{$t = 1$ to $T$}
            \STATE Extract kinematics $\mathbf{v}_{k,t}$, $\mathbf{a}_{k,t}$, and $s_{k,t}$ via Eq.~(9)
            \STATE $\mathbf{z}_{k,t} \leftarrow [\mathbf{y}_{k,t}^{(0)}, \mathbf{v}_{k,t}, \mathbf{a}_{k,t}, s_{k,t}]$
            \STATE $\mathbf{h}_{k,t} \leftarrow \mathrm{MLP}(\mathbf{z}_{k,t}) + \mathbf{c}_{k,t}$
        \ENDFOR
        \STATE $[\boldsymbol{\gamma}_k, \boldsymbol{\beta}_k] \leftarrow \mathrm{MLP}(\boldsymbol{m}_k)$
        \STATE $\bar{\mathbf{h}}_k \leftarrow (1+\tanh(\boldsymbol{\gamma}_k)) \odot \mathbf{h}_k + \boldsymbol{\beta}_k$
        \STATE $\bar{\mathbf{h}}_k \leftarrow \mathrm{SelfAttn}(\bar{\mathbf{h}}_k)$
        \STATE $\mathbf{h}_k' \leftarrow \mathrm{CrossAttn}(\bar{\mathbf{h}}_k, \mathbf{H}_s)$
        \STATE $\Delta \mathbf{Y}_k^{(r)} \leftarrow f_{\mathrm{ref}}(\mathbf{h}_k')$
        \STATE $\mathbf{Y}_k \leftarrow \mathbf{Y}_k^{(0)} + \Delta \mathbf{Y}_k^{(r)}$
    \ENDFOR
    \RETURN $\{\mathbf{Y}_k\}_{k=1}^{K_{\mathrm{macro}}}$
    \end{algorithmic}
    \end{algorithm}
First, explicit kinematic features are extracted from the coarse trajectories and concatenated with the original positions to form enhanced trajectory representations. For the $k$-th trajectory at time step $t$, let $\mathbf{y}_{k,t}$ denote its position. The corresponding velocity, acceleration, and speed are defined as
\begin{equation}
    \begin{aligned}
    \mathbf{v}_{k,t} &= \mathbf{y}_{k,t} - \mathbf{y}_{k,t-1}, \\
    \mathbf{a}_{k,t} &= \mathbf{v}_{k,t} - \mathbf{v}_{k,t-1}, \\
    s_{k,t} &= \|\mathbf{v}_{k,t}\|.
    \end{aligned}
    \end{equation}
The enhanced feature vector is then written as
\begin{equation}
\mathbf{z}_{k,t} = [\mathbf{y}_{k,t}, \mathbf{v}_{k,t}, \mathbf{a}_{k,t}, s_{k,t}].
\end{equation}

These features are projected into a high-dimensional latent space through a multilayer perceptron (MLP) and fused with the hidden states from the decoding stage to form the refinement representation:
\begin{equation}
\mathbf{h}_{k,t} = \operatorname{MLP}(\mathbf{z}_{k,t}) + \mathbf{c}_{k,t},
\end{equation}
where $\mathbf{c}_{k,t}$ denotes the contextual feature inherited from the decoder.

To enable intention-aware refinement, a FiLM-style conditional modulation mechanism is introduced. Specifically, the macro-action embedding $\boldsymbol{m}_k$ corresponding to the $k$-th candidate is used to generate modulation parameters:
\begin{equation}
[\boldsymbol{\gamma}_k, \boldsymbol{\beta}_k] = \operatorname{MLP}(\boldsymbol{m}_k),
\qquad
\boldsymbol{\gamma}_k, \boldsymbol{\beta}_k \in \mathbb{R}^{d_{\text{model}}}.
\end{equation}
A $\tanh$ function is applied to constrain the modulation magnitude, leading to
\begin{equation}
\tilde{\mathbf{h}}_{k,t}
=
\left(1+\tanh(\boldsymbol{\gamma}_k)\right)\odot \mathbf{h}_{k,t}
+
\boldsymbol{\beta}_k,
\end{equation}
where $\odot$ denotes element-wise multiplication.

For sequence modeling, temporal self-attention is first used to capture intra-trajectory dynamic consistency, and cross-attention is then applied to incorporate scene constraints from the global representation:
\begin{equation}
\mathbf{h}'_k
=
\operatorname{CrossAttn}\!\left(
\operatorname{SelfAttn}(\tilde{\mathbf{h}}_k), \mathbf{H}_s
\right).
\end{equation}

Finally, as summarized in Algorithm~\ref{alg:refine}, a residual correction is predicted through a linear mapping and added to the coarse trajectory:
\begin{equation}
    \mathbf{Y}_k = \mathbf{Y}_k^{(0)} + \Delta \mathbf{Y}_k^{(r)}, 
    \qquad
    \Delta \mathbf{Y}_k^{(r)} = f_{\text{ref}}(\mathbf{h}'_k).
\end{equation}

\subsection{Scoring and Ranking}
After obtaining the refined trajectories, we adopt a hybrid evaluation mechanism that combines learning-based and rule-based scoring for candidate ranking. To characterize interaction risk over the prediction horizon, an Interaction World Model is introduced to predict the future trajectories of neighboring vessels:
\begin{equation}
\hat{\mathbf{X}}^{\text{nbr}} = f_w(\mathbf{H}_a, \mathbf{x}_0, \mathbf{v}_0),
\end{equation}
where $\hat{\mathbf{X}}^{\text{nbr}} \in \mathbb{R}^{N_n \times T \times d}$ denotes the predicted future trajectories of neighboring vessels.

\begin{figure}[htbp]
    \centering
    \includegraphics[width=1.0\linewidth]{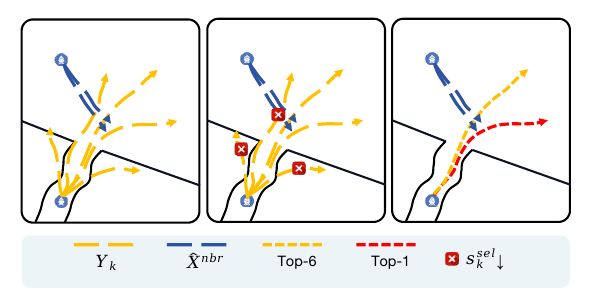}
    \vspace{-15pt}
    \caption{CFR Selector}
    \label{fig:cfr}
\end{figure}

Based on this prediction, a \textbf{Counterfactual Risk (CFR)} assessment mechanism is constructed, as illustrated in Fig.~\ref{fig:cfr}. For each candidate trajectory $\mathbf{Y}_k$, joint analysis is performed with neighboring trajectories to evaluate the interaction risk under the corresponding hypothetical future. Specifically, the distance between the $k$-th candidate trajectory and the $j$-th neighboring vessel at time step $t$ is defined as
\begin{equation}
d_{k,j}(t) = \left\| \mathbf{y}_{k,t} - \hat{\mathbf{x}}_{j,t}^{\text{nbr}} \right\|.
\end{equation}

Based on this distance, two key risk terms are defined. The first is the closest point of approach (CPA) constraint:
\begin{equation}
E_k^{\text{cpa}} =
\sum_j
\phi\!\left(
d_{\text{safe}} - \min_t d_{k,j}(t)
\right),
\end{equation}
and the second is the instantaneous collision risk:
\begin{equation}
E_k^{\text{col}} =
\sum_j \sum_t
\phi\!\left(
d_{\text{col}} - d_{k,j}(t)
\right),
\end{equation}
where $\phi(\cdot)=\log(1+\exp(\cdot))$ denotes the softplus function.

Building upon the interaction risk terms, a comprehensive energy function is further introduced to evaluate physical feasibility and environmental consistency:
\begin{equation}
E_k
=
\lambda_1 E_k^{\text{cpa}}
+
\lambda_2 E_k^{\text{col}}
+
\lambda_3 E_k^{\text{dyn}}
+
\lambda_4 E_k^{\text{map}}
+
\lambda_5 E_k^{\text{land}}
+
\lambda_6 E_k^{\text{head}}.
\end{equation}
Here, $E_k^{\text{dyn}}$ and $E_k^{\text{head}}$ measure trajectory smoothness and heading consistency, while $E_k^{\text{map}}$ and $E_k^{\text{land}}$ penalize lane deviation and boundary violation, respectively.

Meanwhile, a learned scoring model is introduced to evaluate trajectory quality from data. This model fuses sequence features and explicit scalar features to produce a selection score $s_k^{\text{sel}}$ for each candidate trajectory:
\begin{equation}
s_k^{\text{sel}} = f_{\text{sel}}(\mathbf{Y}_k, \mathbf{H}_s).
\end{equation}

This score is combined with the prior probability from the decoder and the normalized energy value $\hat{E}_k$ to obtain a pre-ranking score:
\begin{equation}
s_k^{\text{pre}}
=
\alpha s_k^{\text{prior}}
+
\beta s_k^{\text{sel}}
-
\gamma \hat{E}_k.
\end{equation}

A stronger re-ranking model is then applied to further improve ranking accuracy:
\begin{equation}
s_k^{\text{rerank}} = f_{\text{rerank}}(\mathbf{Y}_k, s_k^{\text{pre}}, E_k, \mathbf{H}_s),
\end{equation}
\begin{equation}
s_k = s_k^{\text{pre}} + \delta s_k^{\text{rerank}}.
\end{equation}

The final probability is normalized through a softmax function:
\begin{equation}
p_k = \frac{\exp(s_k)}{\sum_{j=1}^{K_{\mathrm{macro}}}\exp(s_j)}.
\end{equation}

The trajectory with the highest score,
\begin{equation}
\mathbf{Y}^{*} = \mathbf{Y}_{k^*}, \qquad
k^* = \arg\max_{k \in \{1,\dots,K_{\mathrm{macro}}\}} s_k,
\end{equation}
is selected as the final top-1 prediction, while the top $K_{\mathrm{out}}=6$ trajectories are retained as the multimodal prediction outputs.

\section{Experiments}

\subsection{NaviAIS Dataset and Experimental Setup Analysis}
\begin{figure*}[!t]
    \centering
    \includegraphics[width=1.0\linewidth]{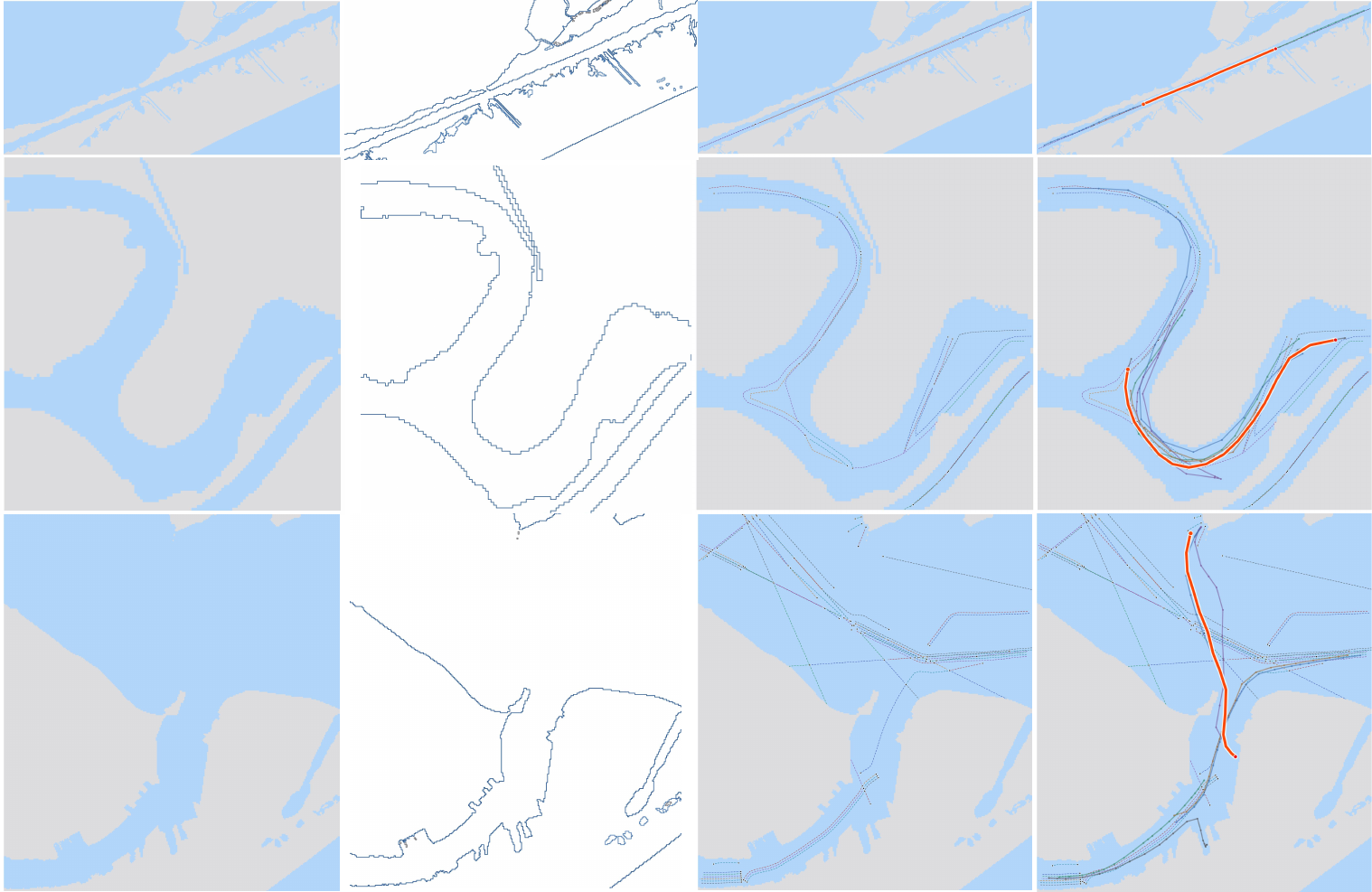}
    \caption{Visualization of dataset scenarios. First show raster maps, vector maps, navigable-channel vectors, and AIS vessel trajectories for narrow inland waterways, respectively. Second show the corresponding results for curved inland waterways. Third show the corresponding results for port areas.}
    \label{fig:datakeshihua_combined}
\end{figure*}
All experiments are conducted on the proposed \textit{NaviAIS} dataset, and the forecasting model is referred to as \textit{NaviLane}. Unlike public AIS resources released as raw message streams, NaviAIS adopts a standardized scenario-level format, where multi-vessel historical--future trajectories, rasterized waterway maps, and vectorized lane priors are jointly stored. The raw source is \textit{USCG AIS Vessel Tracks}, which provides large-scale unprocessed AIS point data and route-related information. In our pipeline, AIS records from September to December 2024 were used to compute aggregated navigational priors, including lane density, route patterns, and waterway attributes. These statistics serve only as static map-related priors and are not used as prediction targets. The scenario-level samples were constructed from representative AIS records in January 2024: the training set uses data from January 1 and contains 4{,}000 scenarios; the validation set uses data from January 4 and contains 1{,}000 scenarios; and the test set uses data from January 7 and also contains 1{,}000 scenarios.

NaviAIS is constructed through a systematic pipeline rather than directly from raw AIS point sequences. First, raw trajectories are cleaned \cite{chen2020ship}, temporally ordered, and filtered to remove abnormal observations \cite{liang2024aisclean}, after which fixed historical--future windows are used to extract candidate clips \cite{zhao2018ship,chen2020ship,liu2024marine}. Second, each scenario is built around a motion-valid target vessel, and both the target and neighboring vessels are transformed into an ego-centric local coordinate system. Third, OSM waterways are used as the global backbone \cite{guo2024unsupervised}, while NOAA line-type features are fused through endpoint snapping, projection, and edge splitting. NOAA area-type elements are further converted into additional lane geometries to preserve wide fairways and route bands \cite{yan2025reconstructing}. Finally, lane centerlines, lane graphs, and optional density maps are queried from offline tiles and injected into each scene according to its spatial extent.

The original route annotations are derived from NOAA ENC electronic chart data. From the \textit{Harbor}, \textit{Approach}, and \textit{Coastal} layers, we extract Navigation\_line, Recommended\_Track\_line, Recommended\_Route\_Centerline\_line, Ferry\_Route\_line, Berth\_line, Canal\_line, and Tideway\_line. For area layers, centerlines are extracted and resampled into multiple parallel route lines to better cover constrained navigation regions.

Each NaviAIS scene contains metadata, trajectory information, raster maps, and vector maps. The metadata include the scenario identifier, temporal range, and participating vessels. The trajectory part contains state tensors, validity masks, vessel identities, ego MMSI, timestamps, temporal interval, and local-coordinate references. The raster-map part includes water masks, land masks, lane-density maps, and raster metadata. The vector-map part includes water polygons, land polygons, lane centerlines, lane attributes, and an explicit lane graph with node coordinates, edge connectivity, and edge polylines \cite{liang2020learning}. This unified representation supports sequence-based forecasting, environment-constrained prediction, and topology-aware reasoning.

\begin{figure}[!t]
    \centering
    \subfloat[]{\includegraphics[width=0.4\linewidth]{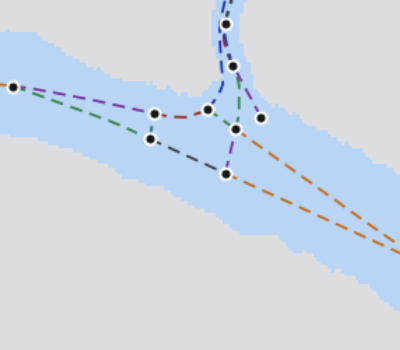}\label{fig:diankeshihua1}}
    \hfil
    \subfloat[]{\includegraphics[width=0.4\linewidth]{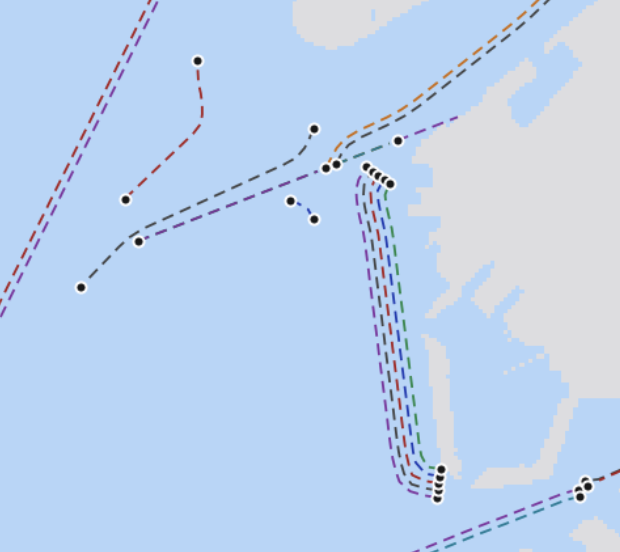}\label{fig:diankeshihua2}}
    \caption{Visualization of local navigable channels.}
    \vspace*{-0.5cm}
    \label{fig:diankeshihua}
\end{figure}

Fig.~\ref{fig:datakeshihua_combined} and Fig.~\ref{fig:diankeshihua} illustrate representative NaviAIS scenarios, including narrow inland waterways, port areas, and curved inland channels. These examples show that raster maps provide coarse water--land occupancy priors, vector maps capture boundary geometry, and lane vectors highlight principal navigational structures. By jointly providing raster and vector representations, NaviAIS enables models to leverage both global occupancy constraints and structured lane-aware priors \cite{gao2020vectornet,xu2022pretram}.

NaviAIS also exhibits strong scene diversity. As shown in Fig.~\ref{fig:datakeshihua_combined}, it covers narrow channels, harbor entrances, and highly curved waterways, requiring models to handle straight-going motion, turning behavior, constrained navigation, and local interaction under different waterway geometries. This makes the benchmark more representative of real-world maritime navigation.

Table~\ref{tab:dataset_comparison} compares NaviAIS with existing vessel trajectory prediction datasets in terms of scale, vectorized lanes, multi-scenario coverage, vectorized maps, openness, and trajectory processing. Most public AIS datasets are large-scale and multi-scenario, but lack vectorized lane priors, structured vector maps, or task-oriented processed trajectories. In contrast, NaviAIS jointly provides scenario-level organization, vectorized lanes, vectorized maps, open accessibility, and processed trajectories, making it a more complete benchmark for environment-aware vessel trajectory prediction.

The proposed model \textit{NaviLane} was trained on NVIDIA A100 GPUs. All baseline and comparison methods were trained and evaluated under the same protocol, using 30 epochs and a batch size of 16. For open-source methods such as DI-MTP and TPTrans, we retained their default configurations and modified only the data loading interface for NaviAIS. All ADE/FDE-related displacement metrics are reported in units of 100 meters, i.e., a value of 1.0 corresponds to 100 meters. For fair multimodal comparison, all methods were constrained to output the best top-6 trajectories, and minADE@6 and minFDE@6 were computed accordingly.

\subsection{Evaluation Metrics}

We evaluate trajectory prediction performance using ADE, FDE, minADE@$K_{\mathrm{out}}$, minFDE@$K_{\mathrm{out}}$, RMSE, MaxDE, and MHE, as defined in Eqs.~\eqref{eq:ade}--\eqref{eq:mhe}. Here, ADE measures the average displacement error over the whole predicted trajectory, while FDE measures the displacement error at the final prediction step. For multimodal prediction, minADE and minFDE denote the minimum ADE and FDE among the top-$K_{\mathrm{out}}$ predicted trajectories\cite{phan2020covernet,zhao2021tnt}, respectively; in this work, we set $K_{\mathrm{out}}=6$. RMSE evaluates the root mean square positional error over the entire trajectory, and MaxDE characterizes the maximum displacement error at the worst time step.

\subsection{Metric Formulations}

Let the ground-truth trajectory be $\mathbf{P}=\{\mathbf{p}_t\}_{t=1}^{T}$ and the $k$-th predicted trajectory be $\hat{\mathbf{P}}^{(k)}=\{\hat{\mathbf{p}}^{(k)}_t\}_{t=1}^{T}$, where $\mathbf{p}_t=(x_t,y_t)\in\mathbb{R}^{2}$ and $\hat{\mathbf{p}}^{(k)}_t=(\hat{x}^{(k)}_t,\hat{y}^{(k)}_t)\in\mathbb{R}^{2}$ denote the ground-truth and predicted 2D positions at time step $t$, respectively.

\begin{equation}
\mathrm{ADE} = \frac{1}{T}\sum_{t=1}^{T} \left\| \hat{\mathbf{p}}_t - \mathbf{p}_t \right\|_2
\label{eq:ade}
\end{equation}

\begin{equation}
\mathrm{FDE} = \left\| \hat{\mathbf{p}}_{T} - \mathbf{p}_{T} \right\|_2
\label{eq:fde}
\end{equation}

\begin{equation}
\mathrm{minADE} = \min_{k \in \{1,\dots,K_{\mathrm{out}}\}} \frac{1}{T}\sum_{t=1}^{T} \left\| \hat{\mathbf{p}}^{(k)}_t - \mathbf{p}_t \right\|_2
\label{eq:minade}
\end{equation}

\begin{equation}
\mathrm{minFDE} = \min_{k \in \{1,\dots,K_{\mathrm{out}}\}} \left\| \hat{\mathbf{p}}^{(k)}_{T} - \mathbf{p}_{T} \right\|_2
\label{eq:minfde}
\end{equation}

\begin{equation}
\mathrm{RMSE} = \sqrt{\frac{1}{T}\sum_{t=1}^{T} \left\| \hat{\mathbf{p}}_t - \mathbf{p}_t \right\|_2^2}
\label{eq:rmse}
\end{equation}

\begin{equation}
\mathrm{MaxDE} = \max_{t \in \{1,\dots,T\}} \left\| \hat{\mathbf{p}}_t - \mathbf{p}_t \right\|_2
\label{eq:maxde}
\end{equation}

\begin{equation}
    \begin{split}
    \theta_t &= \operatorname{atan2}(y_t-y_{t-1},\,x_t-x_{t-1}), \\
    \hat{\theta}_t &= \operatorname{atan2}(\hat{y}_t-\hat{y}_{t-1},\,\hat{x}_t-\hat{x}_{t-1}), \qquad t=2,\dots,T
    \end{split}
    \label{eq:heading}
\end{equation}

\begin{equation}
\mathrm{MHE} = \frac{1}{T-1}\sum_{t=2}^{T}
\left| \operatorname{wrap}\!\left(\hat{\theta}_t-\theta_t\right) \right|
\label{eq:mhe}
\end{equation}

where $\operatorname{wrap}(\Delta\theta)$ maps the angular difference into $(-\pi,\pi]$, which can be written as
\begin{equation}
\operatorname{wrap}(\Delta\theta) = \left((\Delta\theta+\pi)\bmod 2\pi\right)-\pi .
\label{eq:wrap}
\end{equation}

\subsection{Single-Modal Comparison Results}
\begin{table*}[!t]
    \centering
    \caption{Quantitative comparison of different methods on the NaviAIS test set. The best result in each column is shown in bold.}
    \label{tab:single_modal_comparison}
    \setlength{\tabcolsep}{3.5pt}
    \renewcommand{\arraystretch}{1.15}
    \begin{tabular}{lccccccccc}
    \toprule
    Method 
    & ADE@3$\downarrow$ 
    & FDE@3$\downarrow$ 
    & ADE@5$\downarrow$ 
    & FDE@5$\downarrow$ 
    & ADE@10$\downarrow$ 
    & FDE@10$\downarrow$ 
    & RMSE$\downarrow$ 
    & MaxDE$\downarrow$ 
    & MHE$\downarrow$  \\
    \midrule
    LSTM \cite{graves2012long} & \textbf{1.06} & 1.76  & 1.84 & 3.45  & 4.54 & 10.21  & 5.77 & 10.35 & 0.78 \\
    CA \cite{schubert2008comparison}& 1.84 & 3.12 & 3.47 & 6.97  & 9.48 & 22.36  & 11.86 & 22.39 & 0.81 \\
    CV \cite{ammoun2009real}& 1.56 & 2.37  & 2.40 & 4.10  & 4.80 & 9.55  & 5.63 & 9.67 & 0.53 \\
    CTRV \cite{lytrivis2008cooperative}& 1.50 & 2.30  & 2.36 & 4.12  & 4.98 & 10.28  & 5.95 & 10.41 & 0.63 \\
    Transformer \cite{vaswani2017attention} & 2.15 & 3.33  & 3.49 & 6.28 & 7.40 & 14.71  & 8.73 & 14.78 & 1.02 \\
    Tptrans \cite{wang2024tptrans}& 1.47 & 2.23  & 2.30 & 3.98 & 4.60 & 8.85  & 5.38 & 8.99 & 0.53 \\
    Traisformer \cite{nguyen2024transformer}& 1.86 & 2.77  & 2.93 & 5.16 & 6.13 & 12.14  & 7.25 & 12.31 & 0.59 \\
    DI-MTP\cite{zhang2025unified} & 1.80 & 2.74  & 2.82 & 4.91 & 5.56 & 10.63 & 6.47 & 10.77 & 0.72 \\
    PECnet \cite{mangalam2020not}& 1.53 & 2.28  & 2.33 & 3.94 & 4.77 & 9.59 & 5.65 & 9.70 & 0.58 \\
    Ours & 1.18 & \textbf{1.74} & \textbf{1.83}  & \textbf{3.03} & \textbf{3.57} & \textbf{7.03} & \textbf{4.23} & \textbf{7.27} & \textbf{0.44} \\
    \bottomrule
    \end{tabular}
\end{table*}
Table~\ref{tab:single_modal_comparison} reports the quantitative comparison of different methods on the NaviAIS test set. Overall, the proposed \textit{NaviLane} model achieves the best performance on most metrics, especially in medium- and long-horizon prediction \cite{yu2025aisformer}. In particular, NaviLane obtains the best results on FDE@3, ADE@5, FDE@5, ADE@10, FDE@10, RMSE, MaxDE, and MHE. At the longest prediction horizon, NaviLane reduces ADE@10 to 3.57 and FDE@10 to 7.03, while also achieving the lowest RMSE and MaxDE. The best MHE result further shows that NaviLane produces trajectories with more realistic heading evolution and better directional consistency.

A closer examination of the baselines reveals several important patterns. Classical kinematic baselines such as CV, CTRV, and CA remain relatively competitive at short horizons, which is expected because they rely on strong motion priors and can extrapolate local trends in simple scenarios. However, their errors increase rapidly as the prediction horizon grows. This phenomenon is particularly clear for CA, whose ADE@10 and FDE@10 are substantially worse than those of the proposed model. Such degradation suggests that fixed motion assumptions are insufficient for vessel forecasting in realistic waterways, where trajectory evolution is jointly affected by lane geometry, local constraints, and nearby vessel interaction.

Sequence-based learning methods such as LSTM, Transformer, and TrAISformer improve the ability to model temporal dependencies, but they still underperform NaviLane on most medium- and long-horizon metrics. This result indicates that historical trajectory sequences alone are not enough to fully characterize vessel motion in structured navigational environments. In contrast, NaviLane explicitly incorporates vectorized lane priors and scene-level map context, which allows the model to distinguish between geometrically feasible and infeasible futures. This is especially beneficial in narrow inland channels, curved waterways, and constrained port regions.

Compared with stronger learned baselines such as TPTrans, DI-MTP, and PECNet, NaviLane still maintains a clear advantage. This suggests that the gain of the proposed framework does not come from a larger generic backbone alone, but from the combination of several task-specific design choices: trajectory--map joint encoding, strategy-level macro-action organization, local residual refinement, and world-model-based consequence-aware reranking. In other words, NaviLane improves both the representation of the scene and the mechanism used to generate and select future trajectories.

Another notable observation is that NaviLane does not simply optimize endpoint accuracy at the cost of intermediate trajectory quality. The simultaneous improvement in ADE, FDE, RMSE, MaxDE, and MHE indicates that the predicted trajectories are globally closer to the ground truth, locally smoother, and directionally more faithful. Since maritime motion is usually continuous, inertia-dominated, and constrained by waterway structure, a method that reduces average displacement but produces unrealistic directional changes would still be undesirable in practice. The strong MHE performance of NaviLane therefore provides additional evidence that its predictions are not only accurate but also physically and behaviorally plausible.
\subsection{Multimodal Comparison Results}
\begin{table*}[!t]
    \centering
    \caption{Quantitative comparison of different methods in terms of multimodal trajectory prediction metrics on the NaviAIS dataset. The best result in each column is highlighted in bold.}
    \label{tab:multimodal_comparison}
    \setlength{\tabcolsep}{3.5pt}
    \renewcommand{\arraystretch}{1.15}
    \begin{tabular}{lcccccc}
    \toprule
    Method 
    & minADE@3$\downarrow$ 
    & minFDE@3$\downarrow$
    & minADE@5$\downarrow$ 
    & minFDE@5$\downarrow$ 
    & minADE@10$\downarrow$ 
    & minFDE@10$\downarrow$  \\
    \midrule
    DI-MTP \cite{zhang2025unified} & 1.05 & 1.52  & 1.64 & 2.66 & 5.56 & 10.63  \\
    PECnet \cite{mangalam2020not}& 1.53 & 2.28  & 2.33 & 3.94 & 4.77 & 9.59  \\
    Ours & \textbf{0.97} & \textbf{1.43} & \textbf{1.49}  & \textbf{2.44} & \textbf{2.61} & \textbf{4.85}  \\
    \bottomrule
    \end{tabular}
\end{table*}

Table~\ref{tab:multimodal_comparison} further compares different methods in terms of multimodal forecasting metrics. NaviLane achieves the best results on all listed metrics, including minADE@3, minFDE@3, minADE@5, minFDE@5, minADE@10, and minFDE@10. The improvements are particularly evident at the long horizon, where NaviLane reaches 2.61 on minADE@10 and 4.85 on minFDE@10, substantially outperforming DI-MTP and PECNet. These results show that the top-6 candidates generated by NaviLane provide stronger coverage of the true future.

The superiority of NaviLane in the multimodal setting suggests that its candidate set is not only diverse, but also well organized. In many multimodal prediction frameworks, diversity is introduced implicitly through latent sampling or endpoint perturbation, which can lead to redundant modes or overly dispersed predictions. In contrast, NaviLane organizes the multimodal space through a discrete macro-action codebook, explicitly partitioning plausible high-level navigational strategies before decoding continuous trajectories. This design makes different modes more interpretable and reduces redundant candidates.

The refinement and reranking stages further contribute to this advantage. After coarse strategy-level generation, the residual refinement module improves local geometric precision while preserving the global mode structure. The world-model-based counterfactual evaluator then selects candidates not only according to their prior likelihood, but also according to their anticipated interaction consequences. As a result, NaviLane achieves a better balance between diversity and rationality, which is important for maintaining consistency with channel geometry, vessel interaction, and navigational feasibility.

It is also worth noting that the multimodal advantage of NaviLane is aligned with the single-modal results. This indicates that the model improves both candidate generation quality and final ranking, rather than merely relying on a large set of candidates. In practical terms, such an informative and well-structured candidate set is more suitable for downstream decision-making and risk analysis.

\subsection{Qualitative Analysis of Prediction Results}
\begin{figure*}[!t]
    \centering
    \subfloat[]{\includegraphics[width=0.235\textwidth]{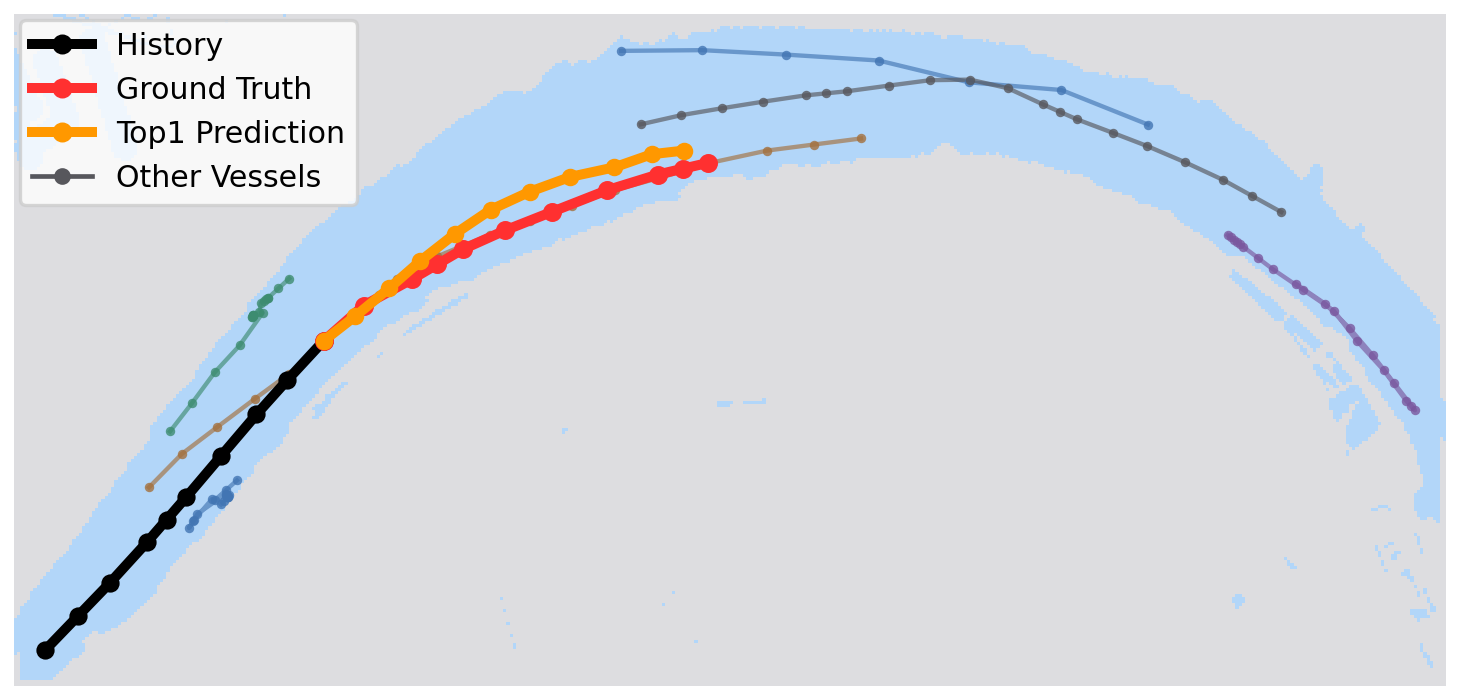}\label{fig:qubiekeshihua11}}
    \hfil
    \subfloat[]{\includegraphics[width=0.235\textwidth]{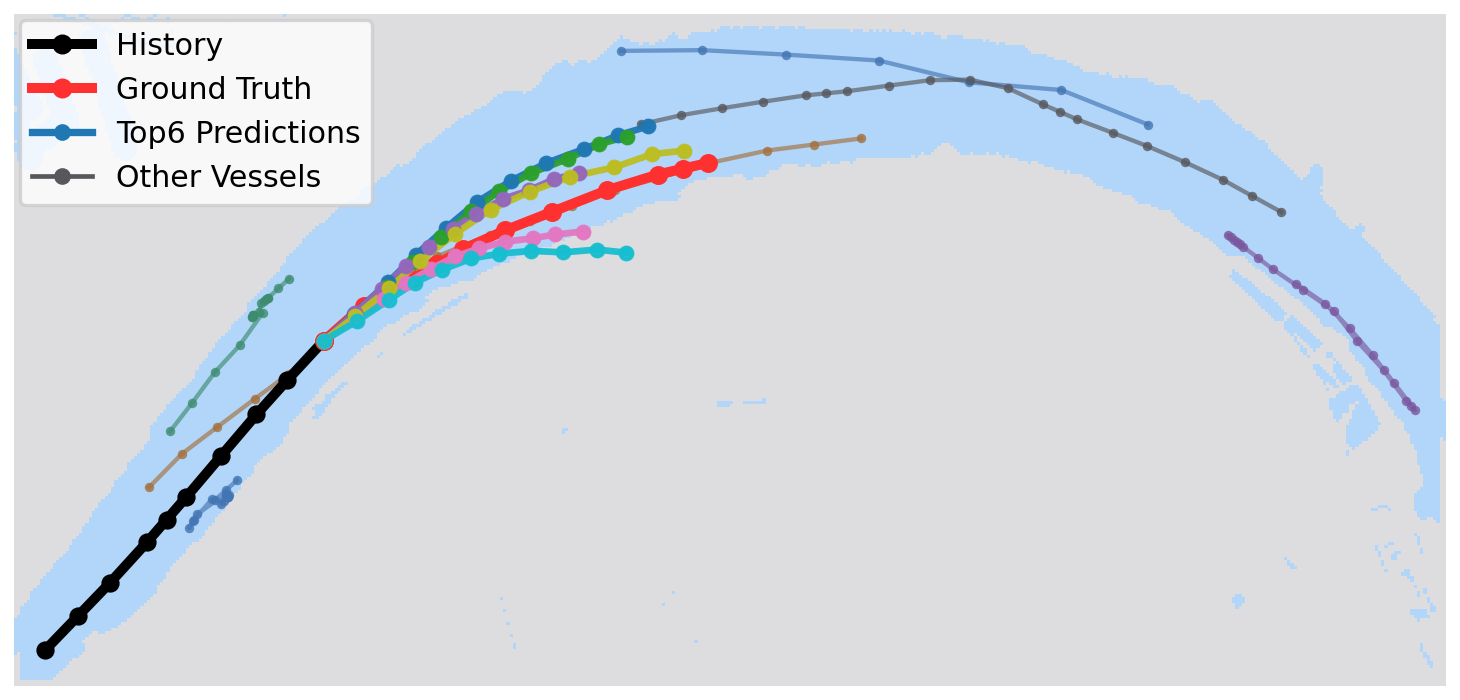}\label{fig:qubiekeshihua12}}
    \hfil
    \subfloat[]{\includegraphics[width=0.235\textwidth]{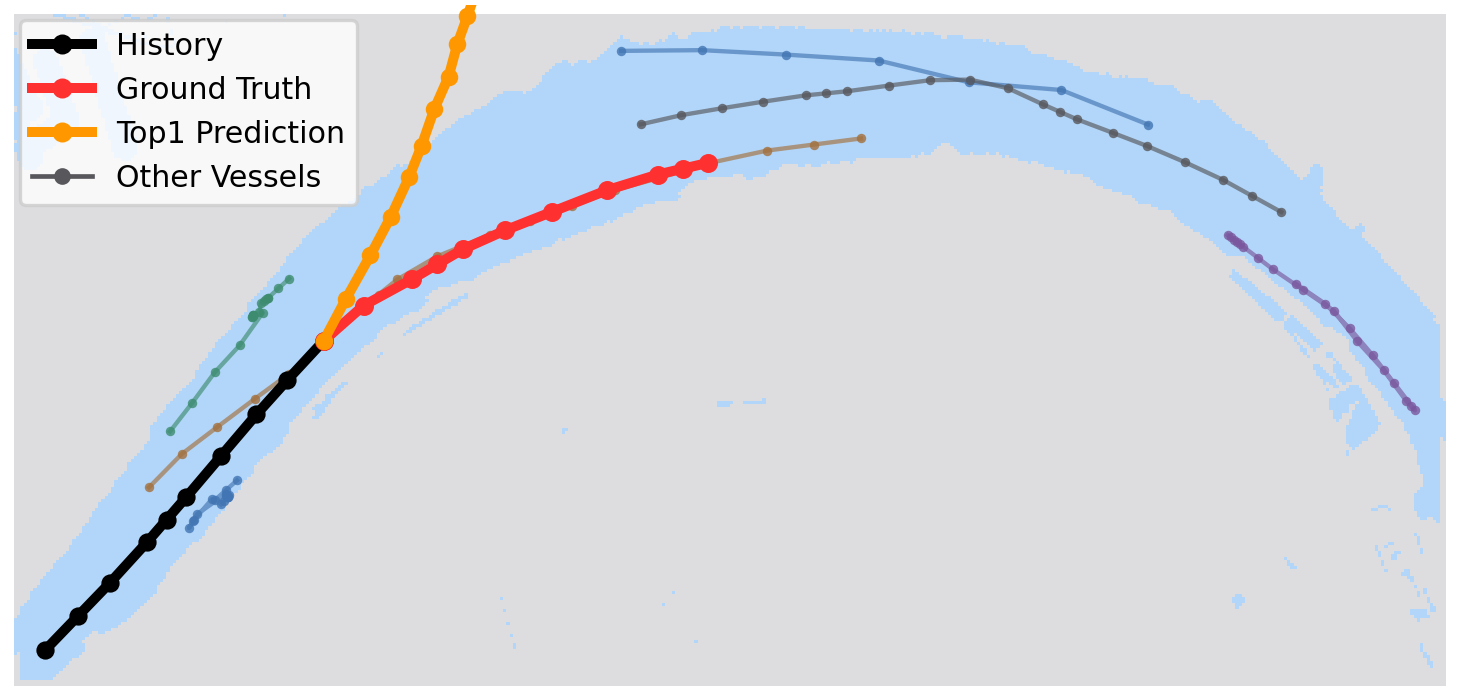}\label{fig:qubiekeshihua13}}
    \hfil
    \subfloat[]{\includegraphics[width=0.235\textwidth]{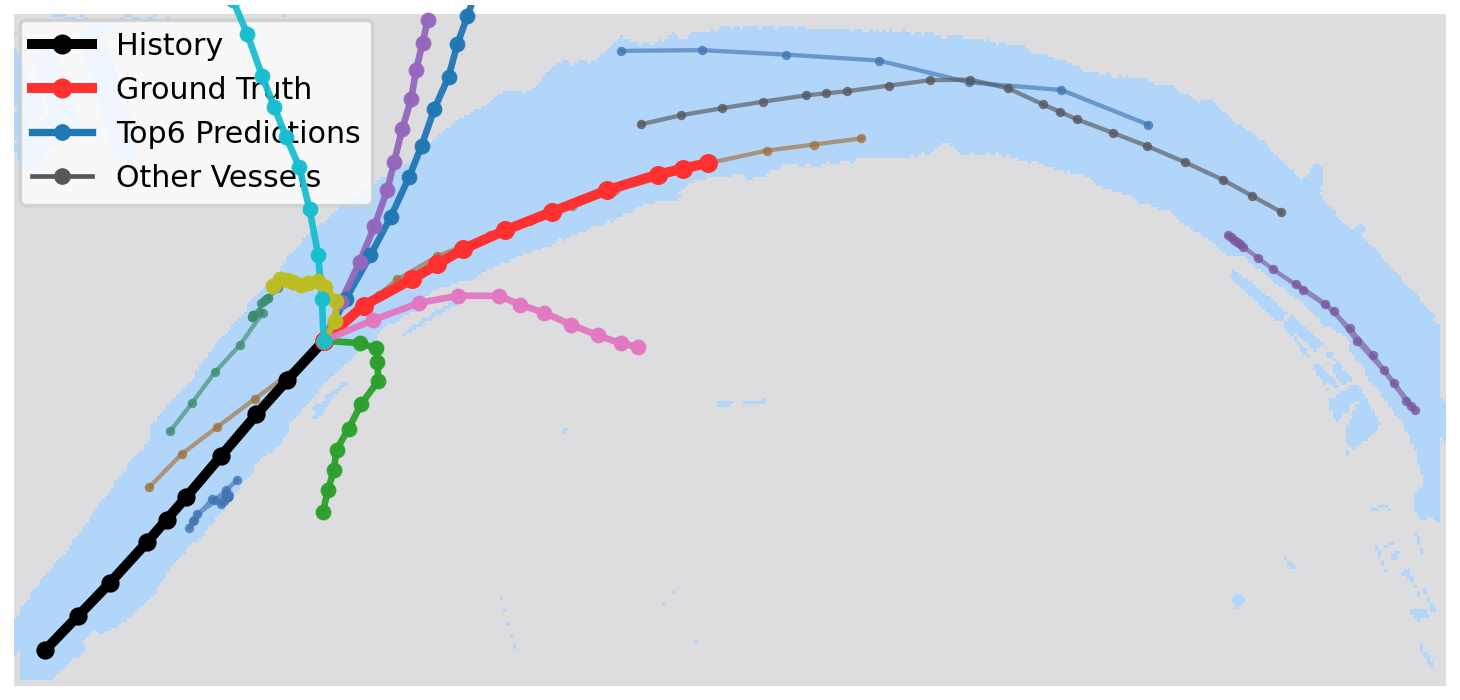}\label{fig:qubiekeshihua14}}
    \hfil
    \subfloat[]{\includegraphics[width=0.235\textwidth]{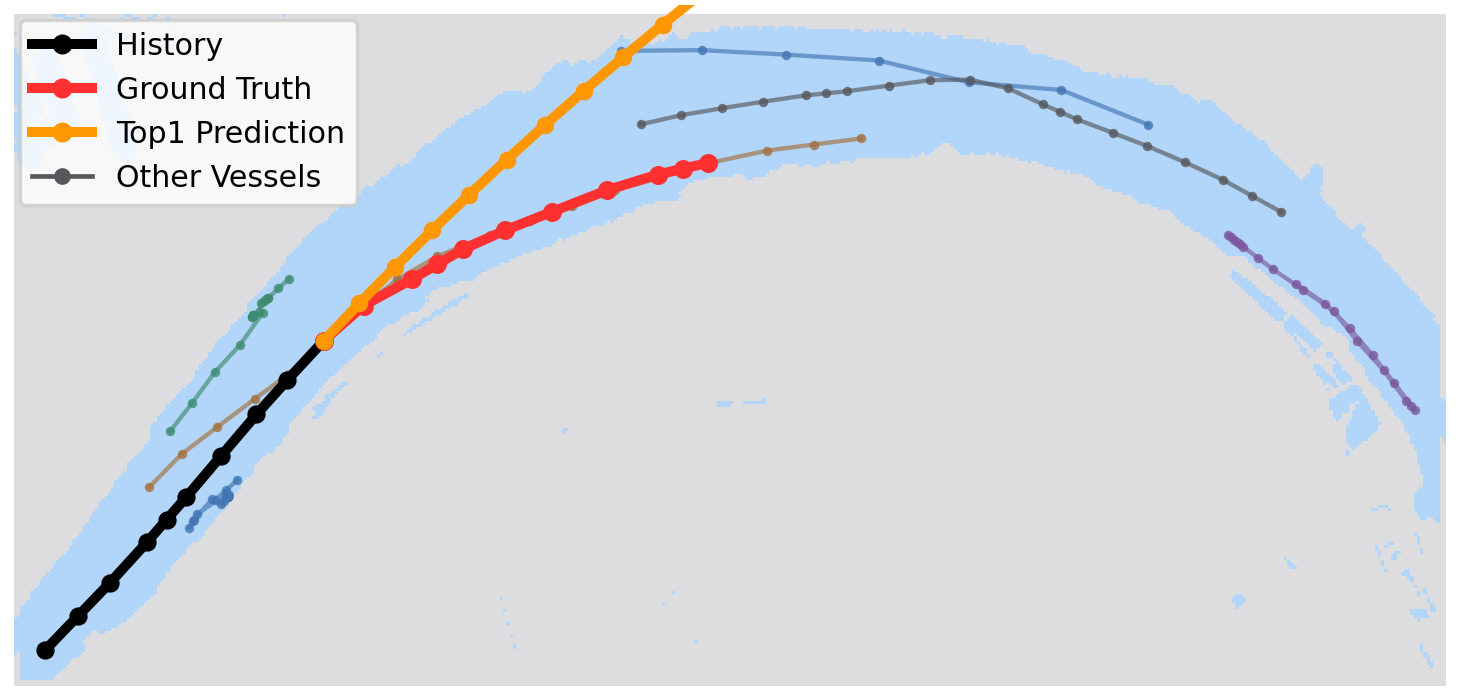}\label{fig:qubiekeshihua15}}
    \hfil
    \subfloat[]{\includegraphics[width=0.235\textwidth]{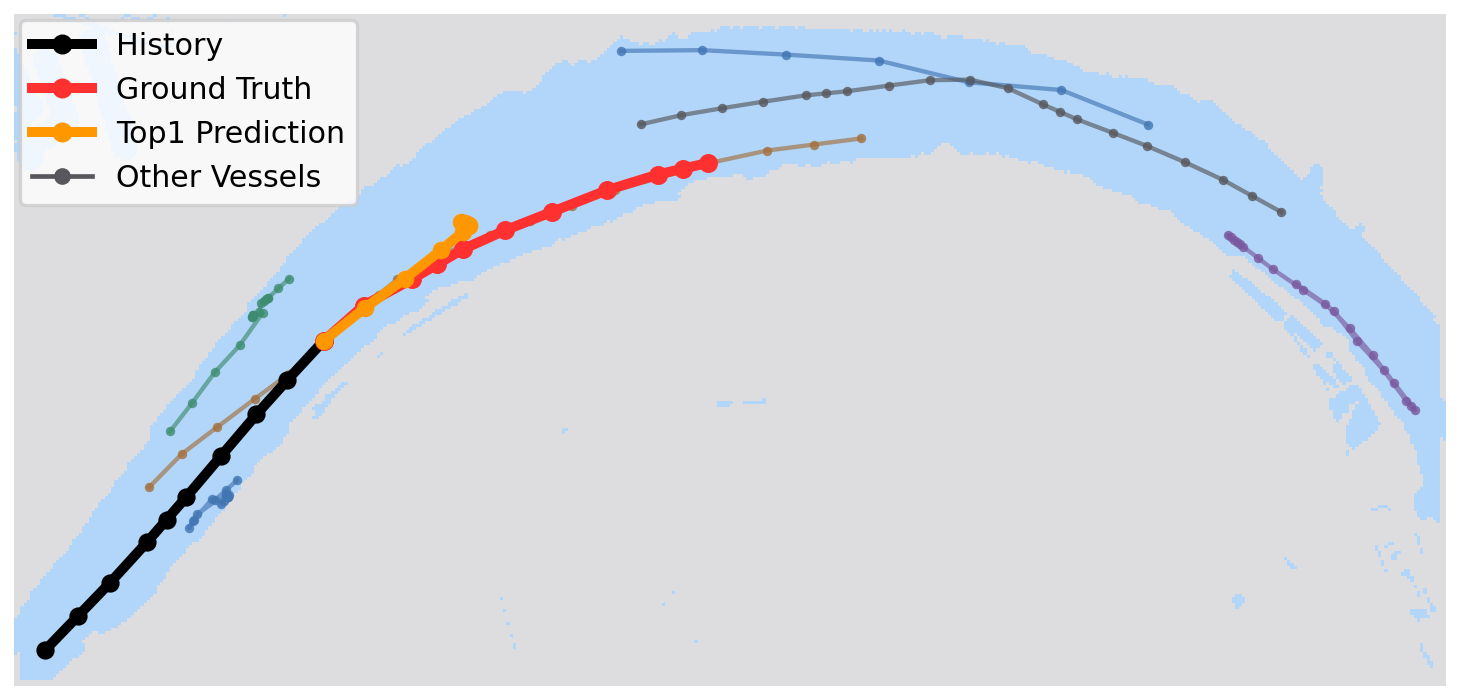}\label{fig:qubiekeshihua16}}
    \hfil
    \subfloat[]{\includegraphics[width=0.235\textwidth]{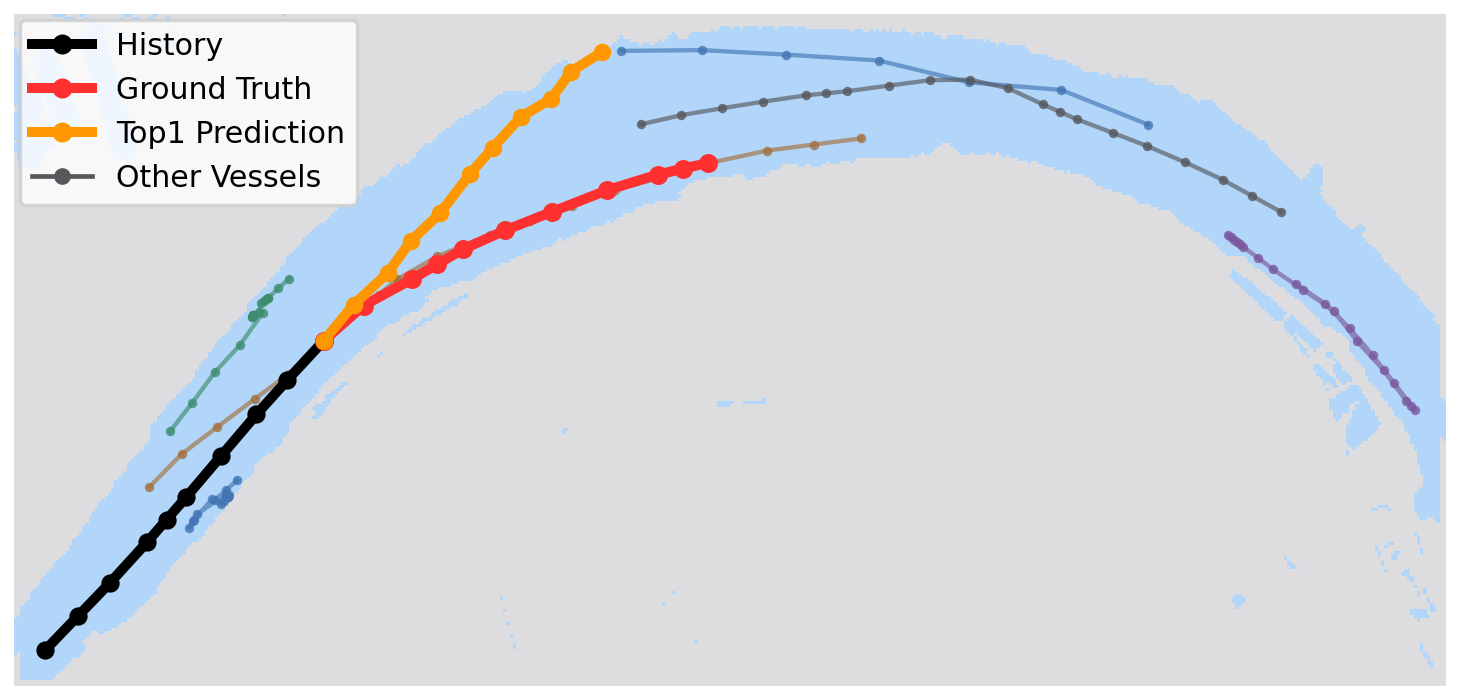}\label{fig:qubiekeshihua17}}
    \hfil
    \subfloat[]{\includegraphics[width=0.235\textwidth]{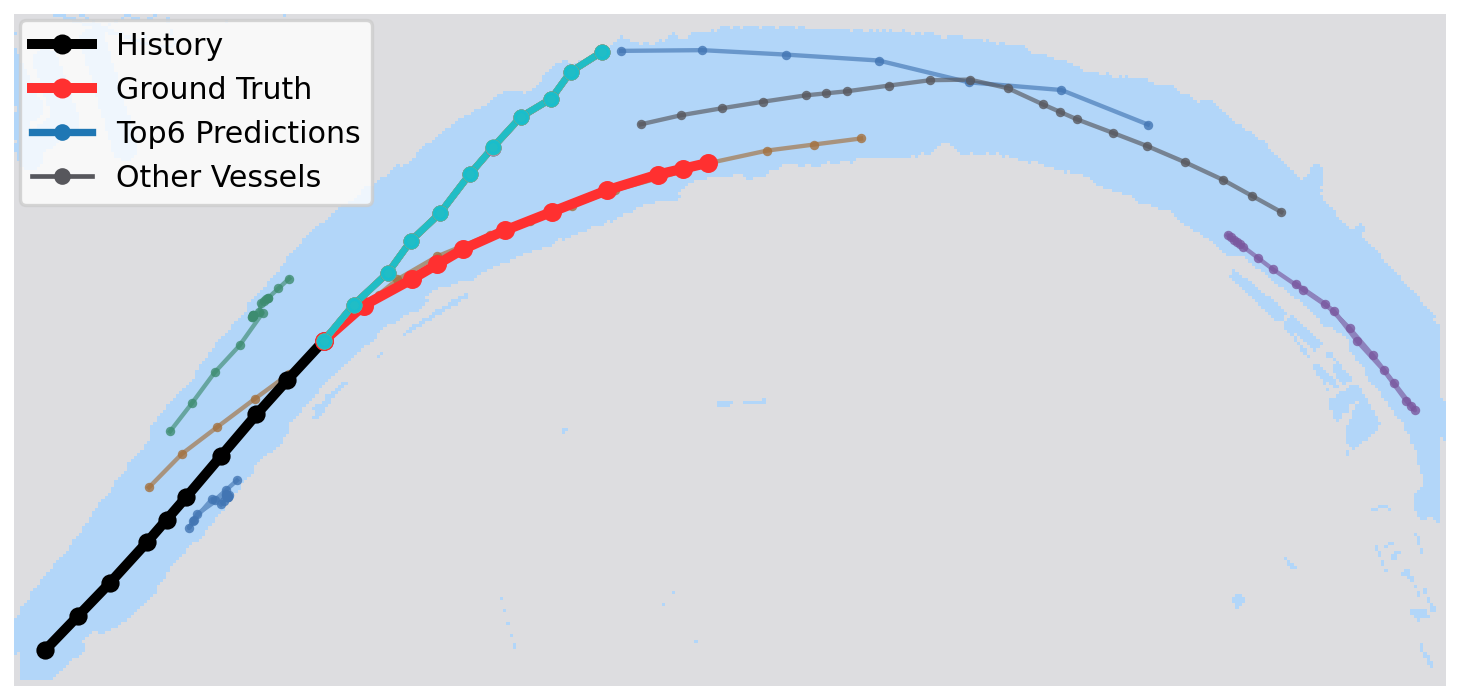}\label{fig:qubiekeshihua18}}
    \hfil
    \subfloat[]{\includegraphics[width=0.235\textwidth]{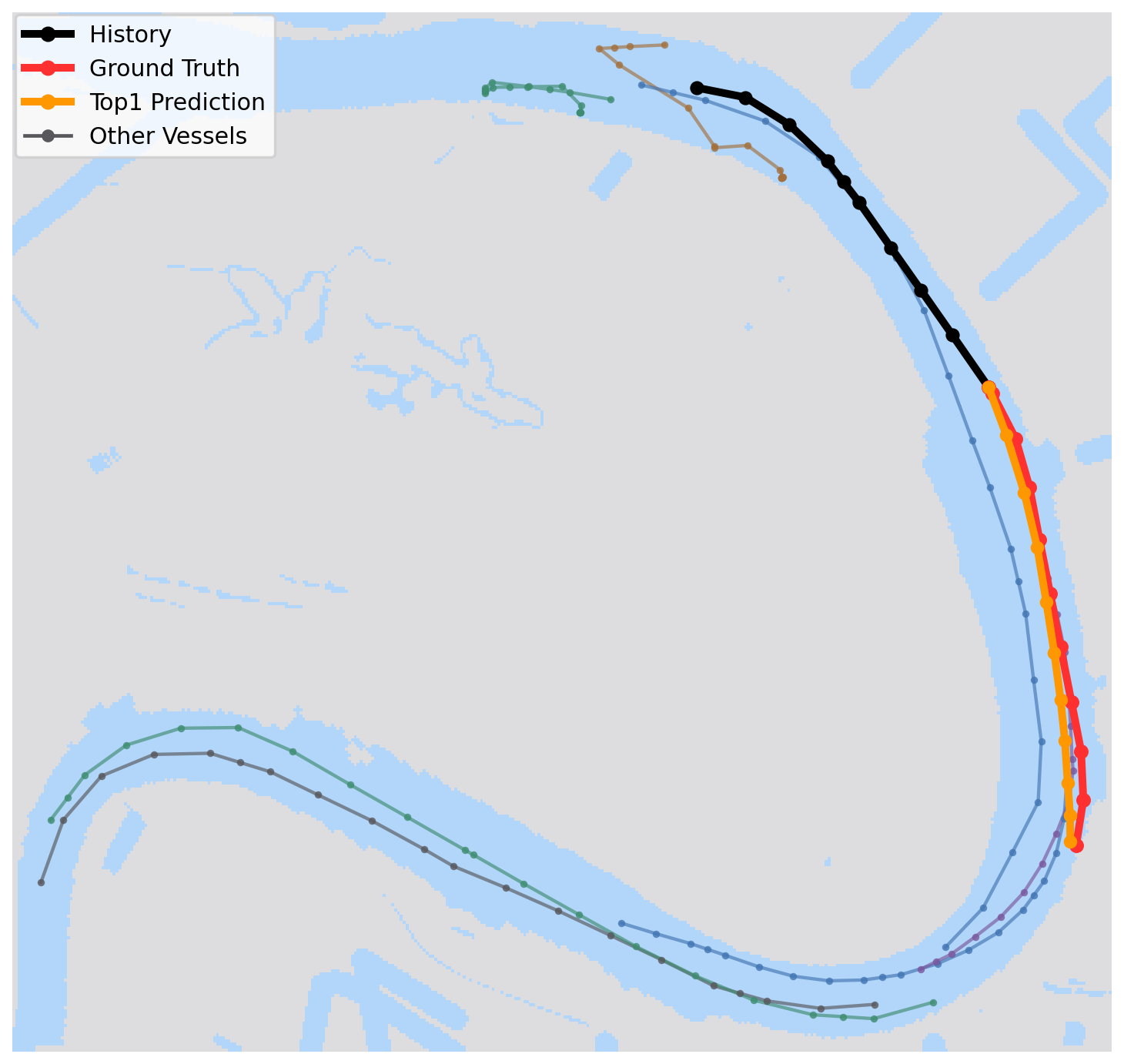}\label{fig:qubiekeshihua21}}
    \hfil
    \subfloat[]{\includegraphics[width=0.235\textwidth]{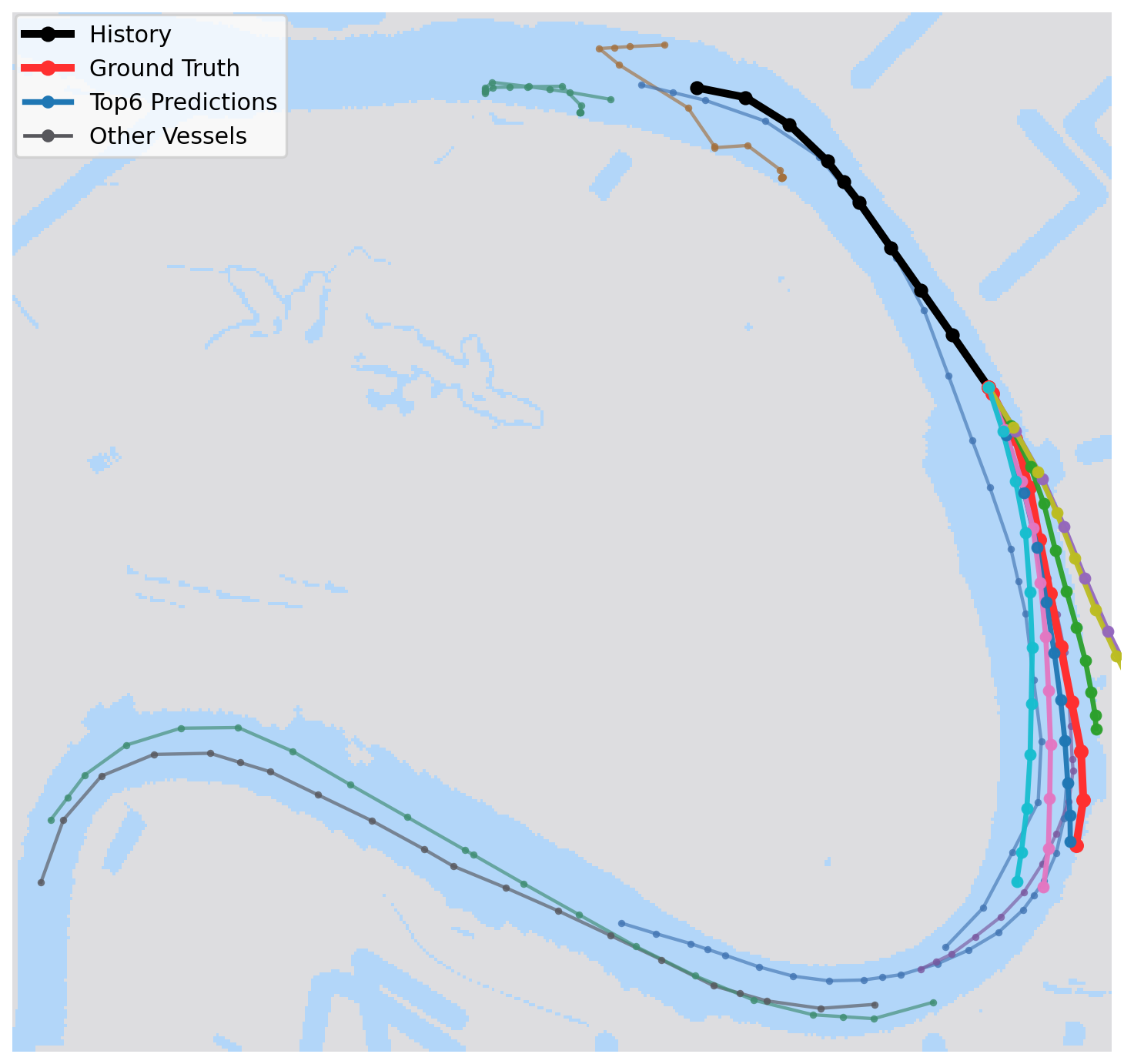}\label{fig:qubiekeshihua22}}
    \hfil
    \subfloat[]{\includegraphics[width=0.235\textwidth]{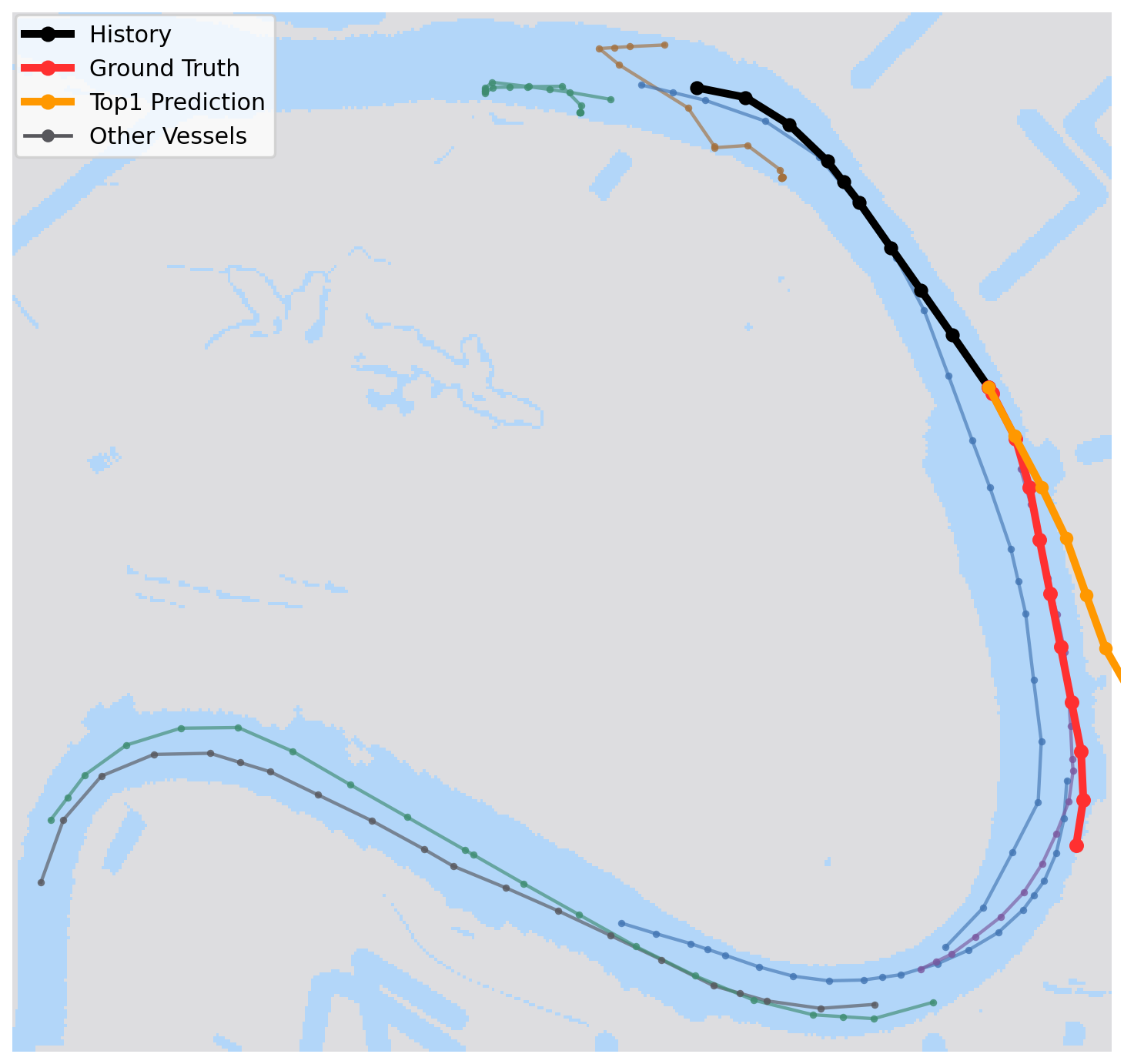}\label{fig:qubiekeshihua23}}
    \hfil
    \subfloat[]{\includegraphics[width=0.235\textwidth]{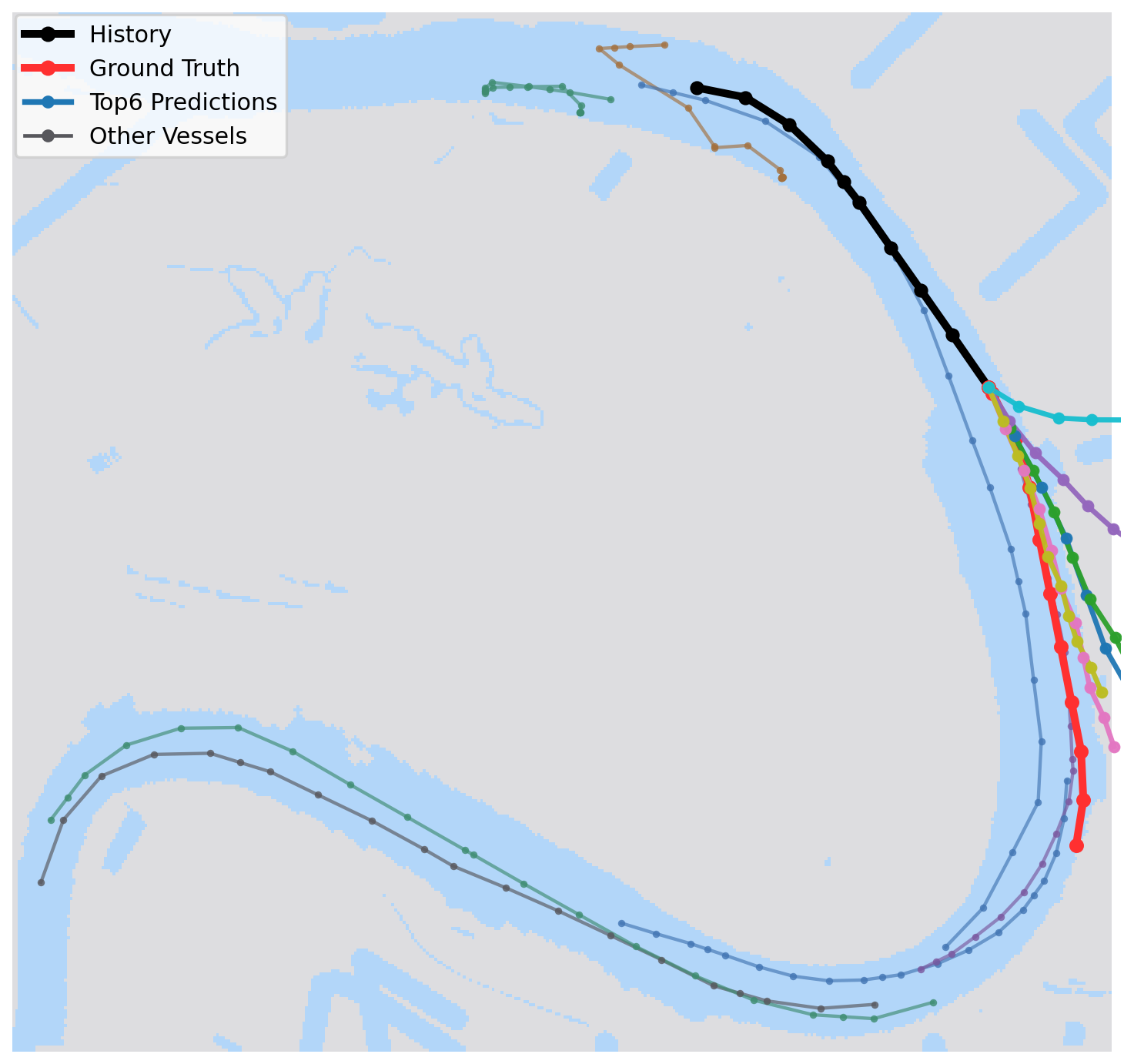}\label{fig:qubiekeshihua24}}
    \hfil
    \subfloat[]{\includegraphics[width=0.235\textwidth]{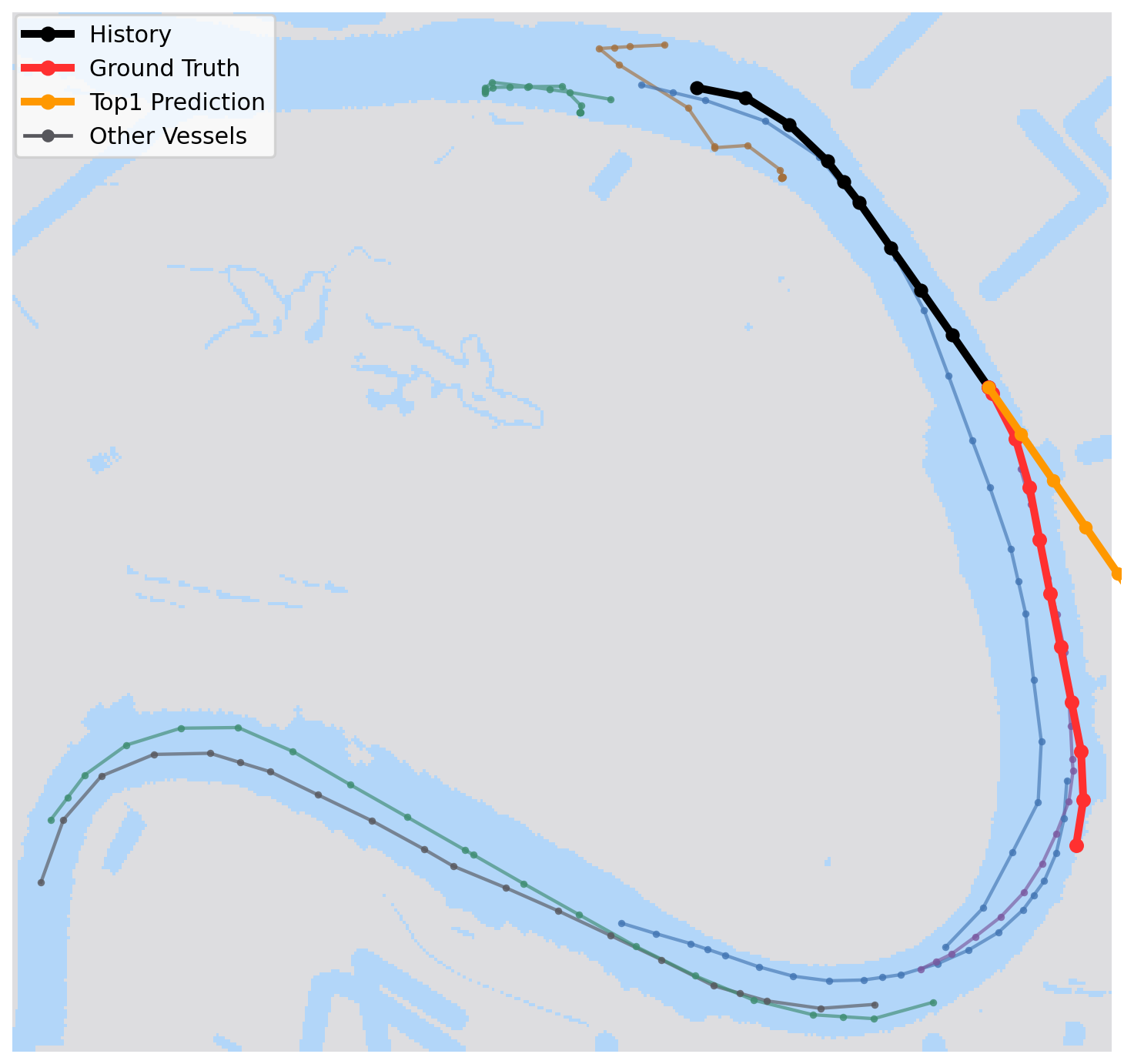}\label{fig:qubiekeshihua25}}
    \hfil
    \subfloat[]{\includegraphics[width=0.235\textwidth]{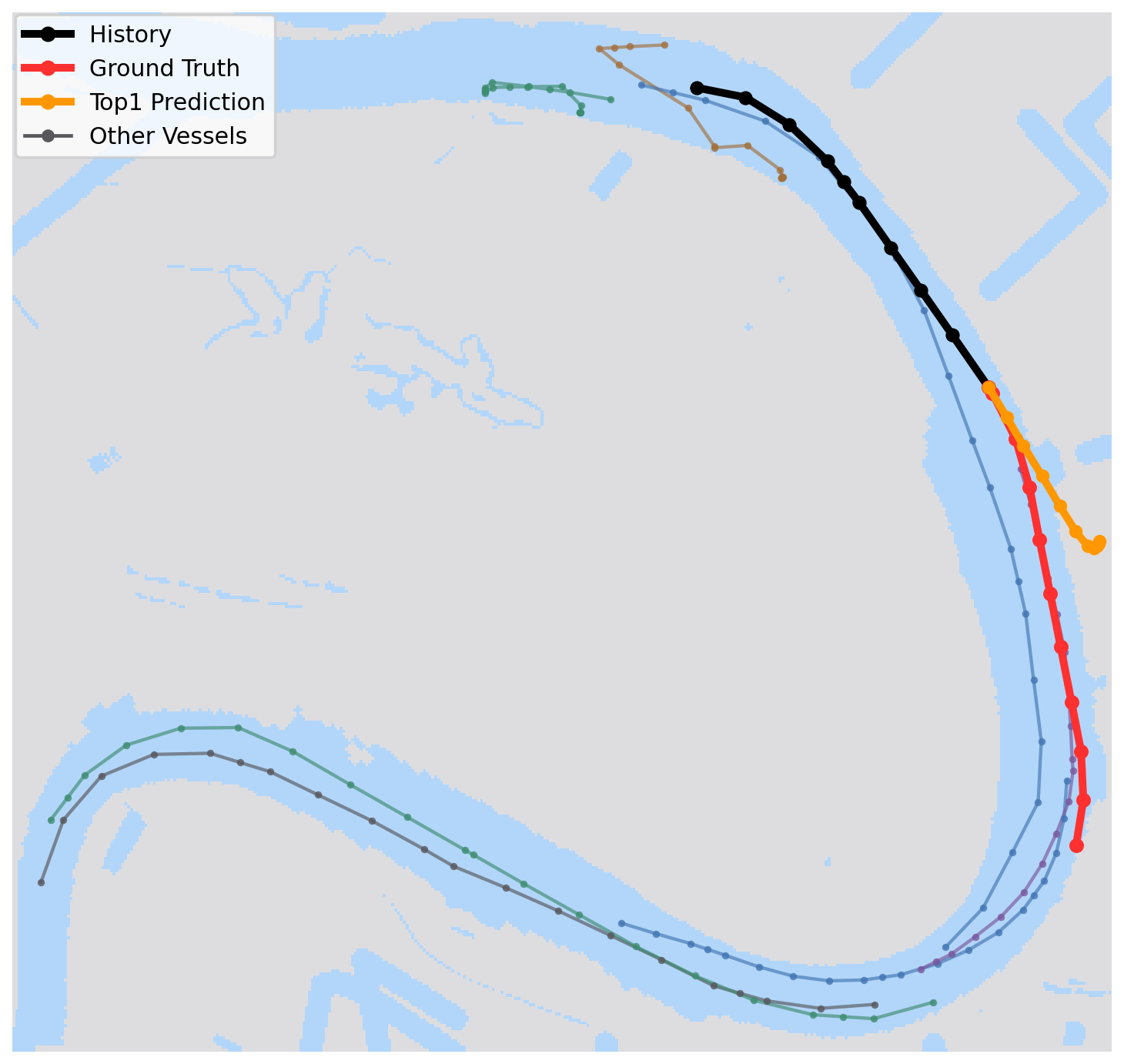}\label{fig:qubiekeshihua26}}
    \hfil
    \subfloat[]{\includegraphics[width=0.235\textwidth]{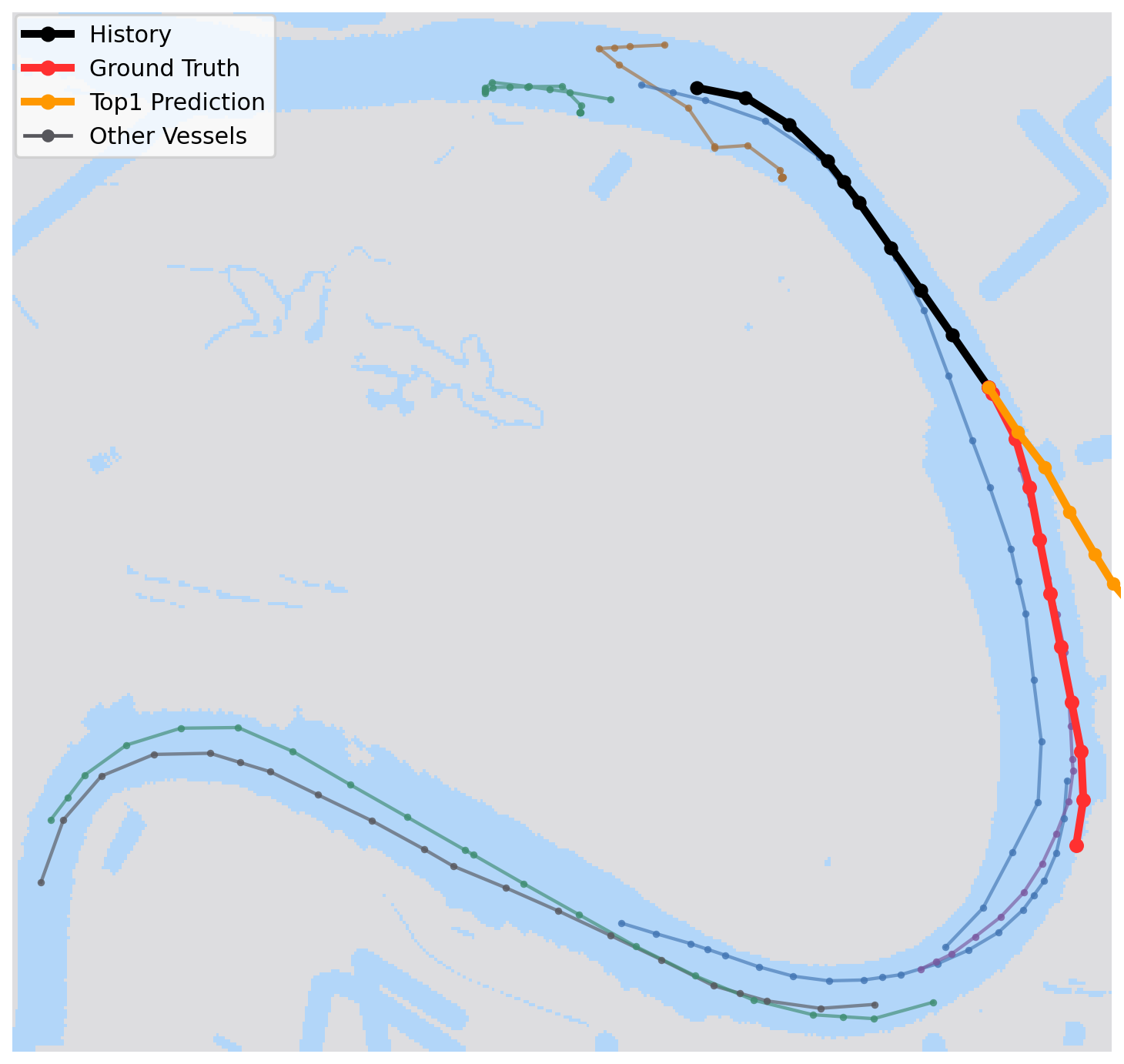}\label{fig:qubiekeshihua27}}
    \hfil
    \subfloat[]{\includegraphics[width=0.235\textwidth]{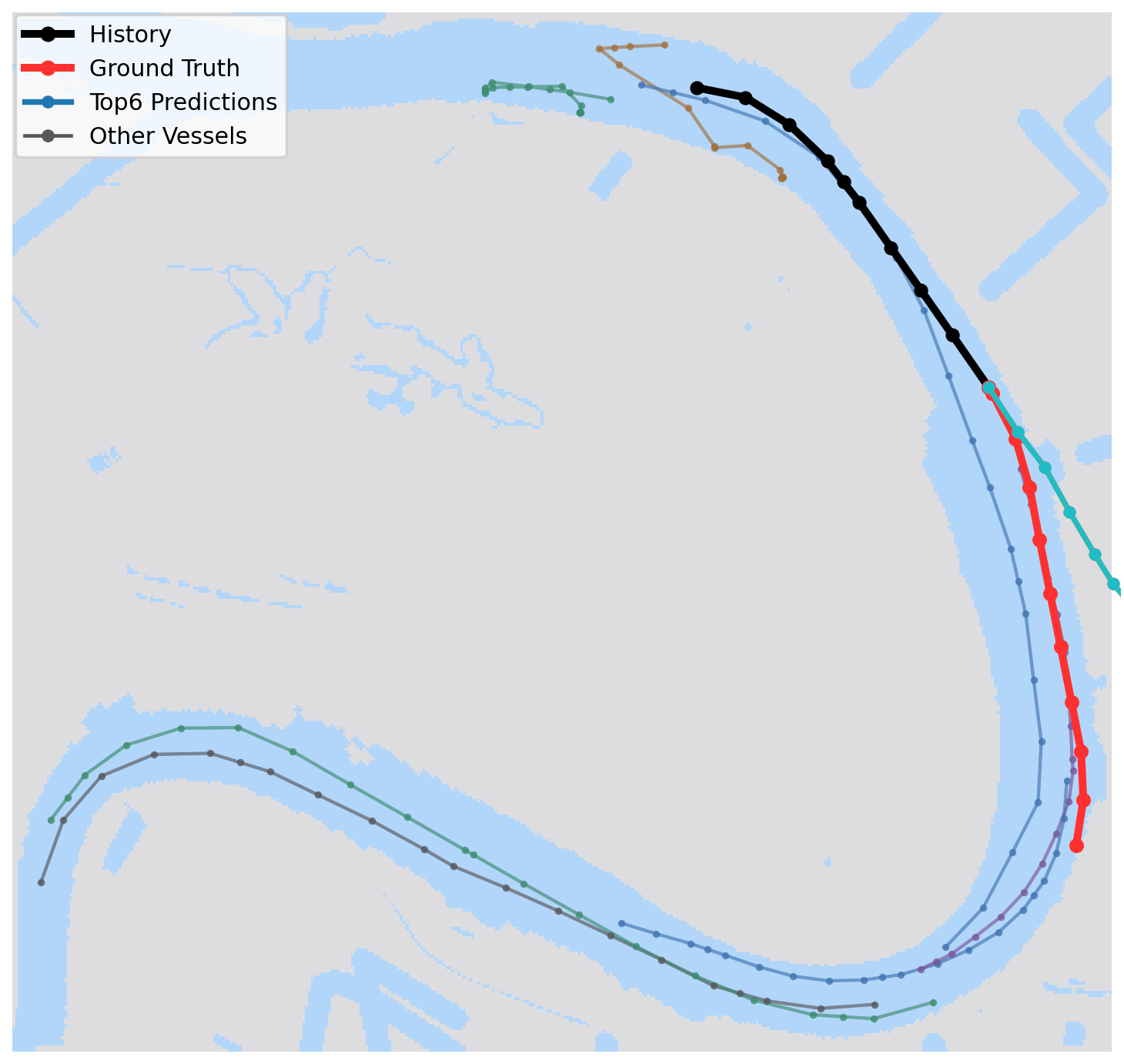}\label{fig:qubiekeshihua28}}
    \caption{Visualization of trajectory prediction results of different methods in inland waterways. (a), (i): NaviLane (top-1); (b), (j): NaviLane (top-6); (c), (k): DI-MTP (top-1); (d), (l): DI-MTP (top-6); (e), (m): CTRV; (f), (n): LSTM; (g), (o): PECNet (top-1); (h), (p): PECNet (top-6).}
    \label{fig:qubiekeshihua2}
\end{figure*}

\begin{figure*}[!t]
    \centering
    \subfloat[]{\includegraphics[width=0.12\textwidth]{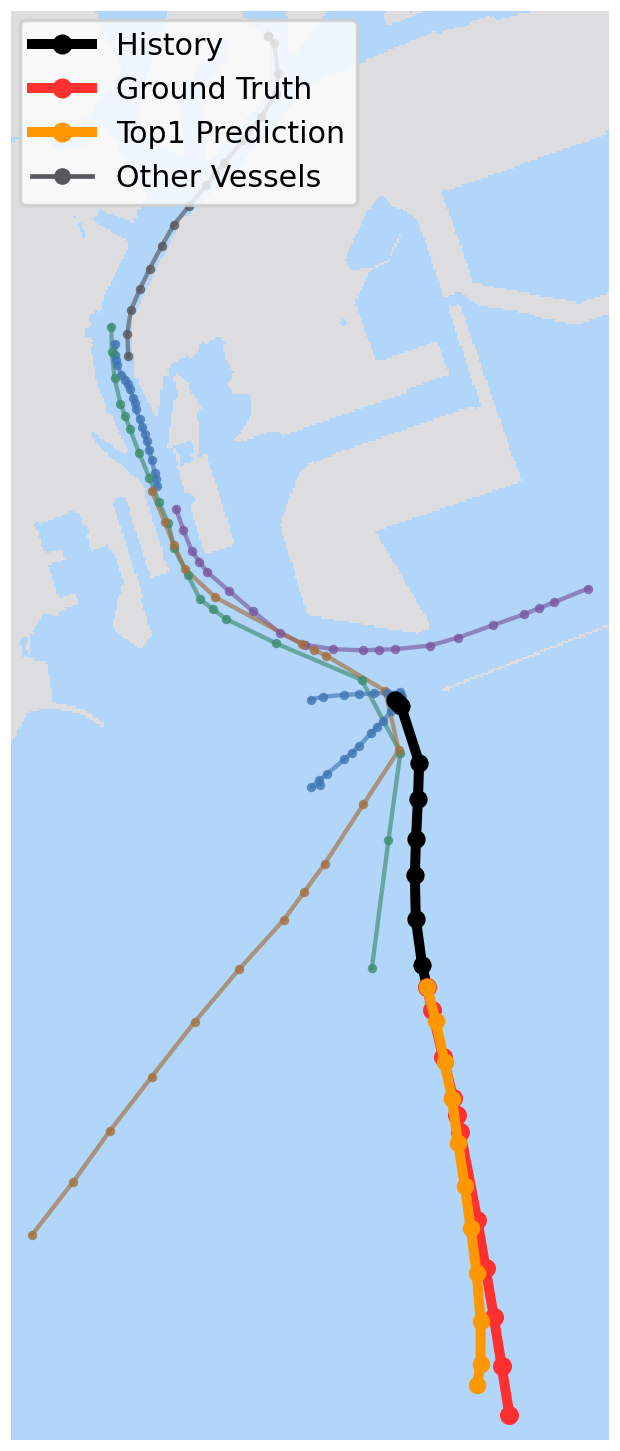}\label{fig:qubiekeshihua31}}
    \hfil
    \subfloat[]{\includegraphics[width=0.12\textwidth]{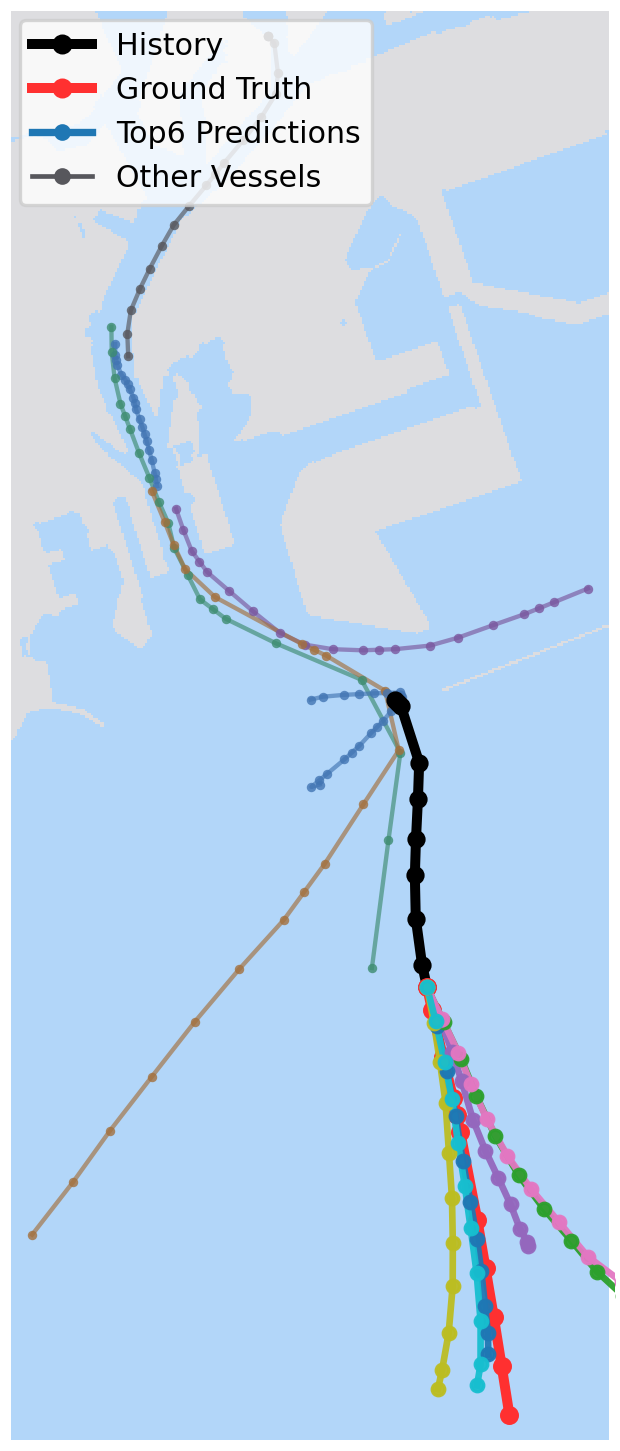}\label{fig:qubiekeshihua32}}
    \hfil
    \subfloat[]{\includegraphics[width=0.12\textwidth]{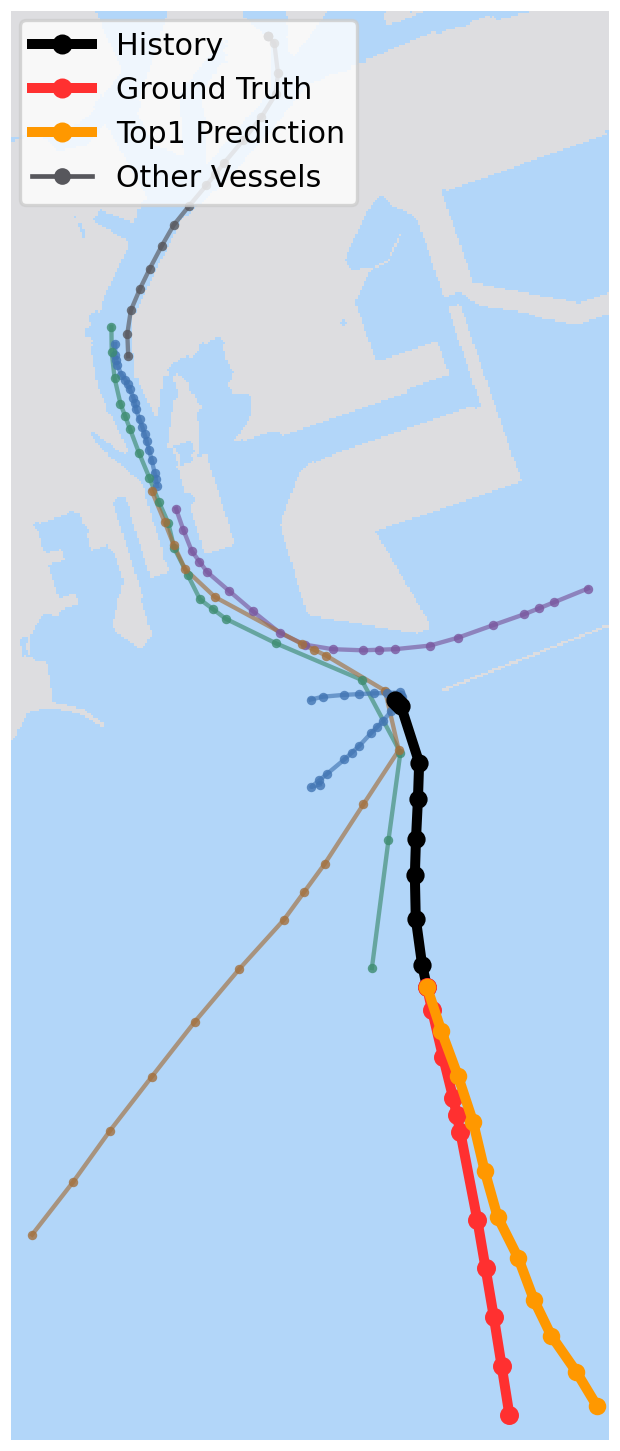}\label{fig:qubiekeshihua33}}
    \hfil
    \subfloat[]{\includegraphics[width=0.12\textwidth]{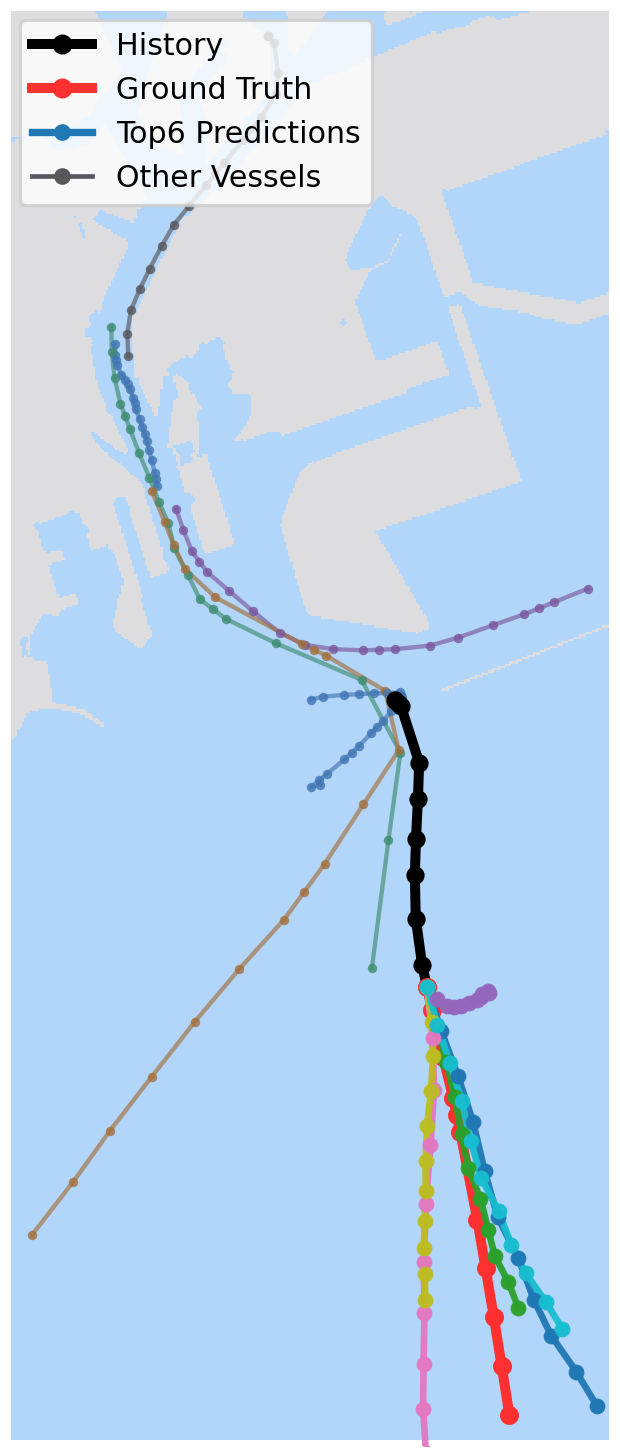}\label{fig:qubiekeshihua34}}
    \hfil
    \subfloat[]{\includegraphics[width=0.12\textwidth]{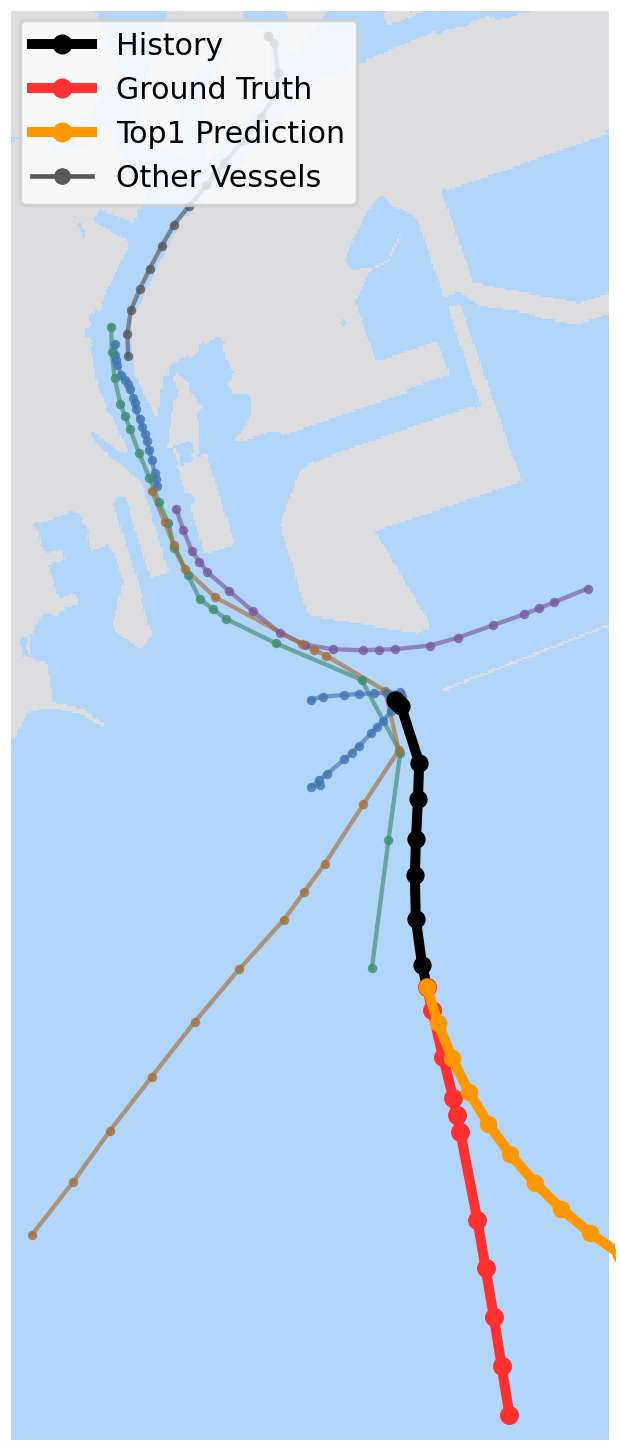}\label{fig:qubiekeshihua35}}
    \hfil
    \subfloat[]{\includegraphics[width=0.12\textwidth]{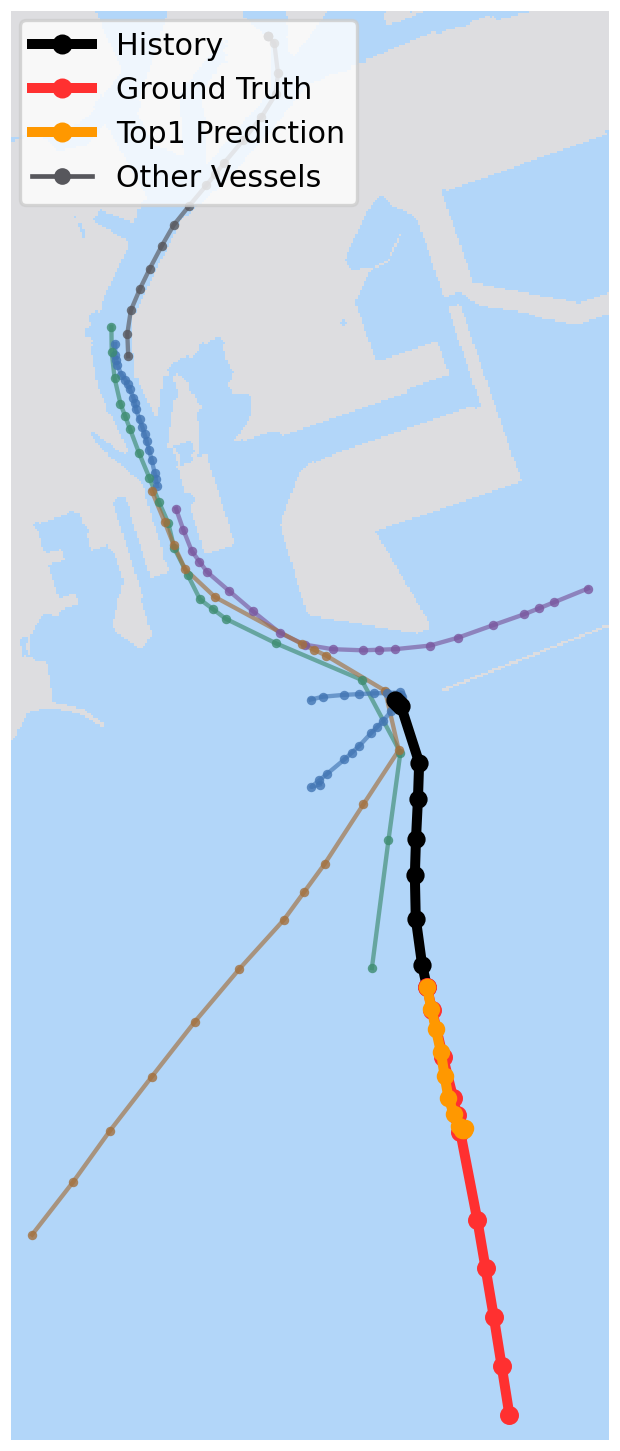}\label{fig:qubiekeshihua36}}
    \hfil
    \subfloat[]{\includegraphics[width=0.12\textwidth]{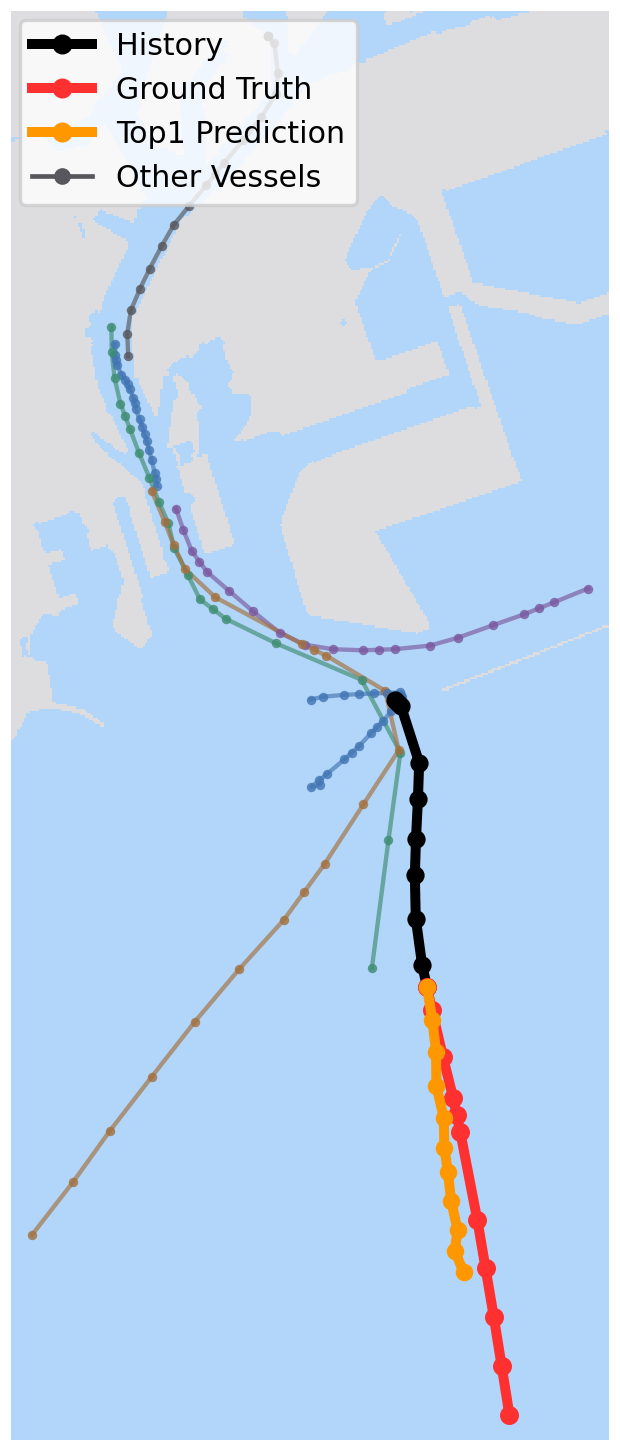}\label{fig:qubiekeshihua37}}
    \hfil
    \subfloat[]{\includegraphics[width=0.12\textwidth]{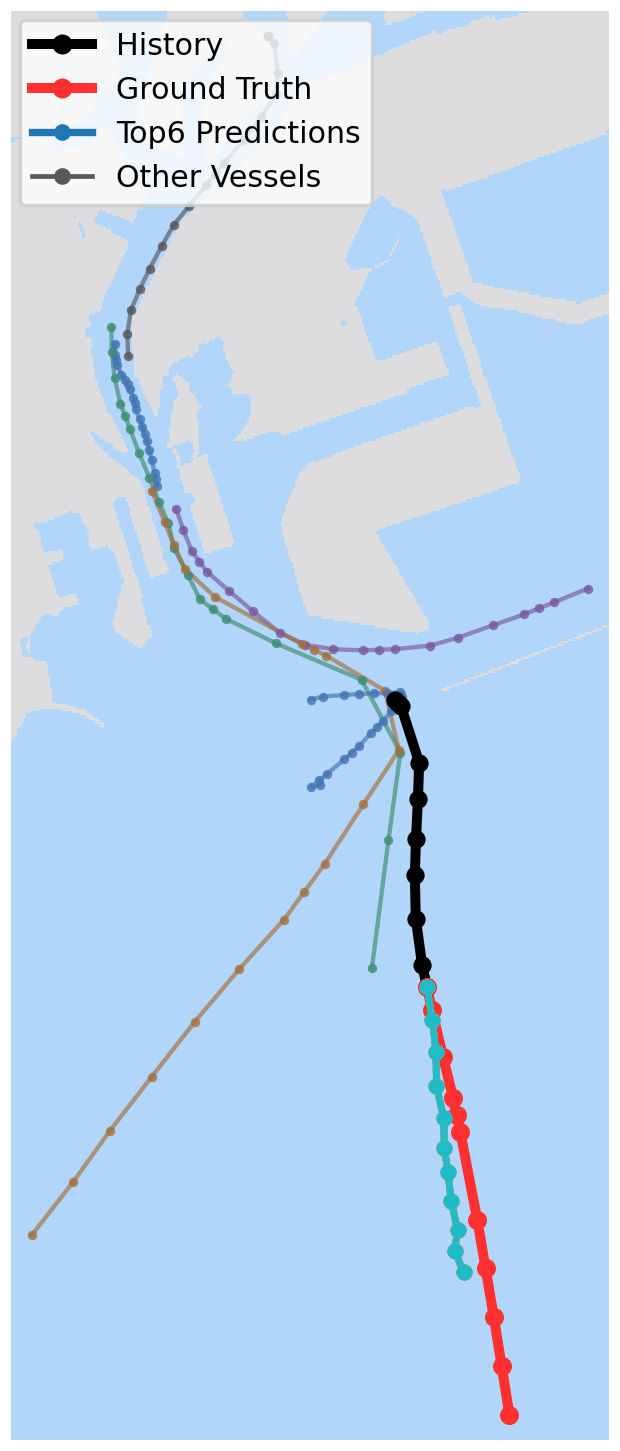}\label{fig:qubiekeshihua38}}
    \hfil
    \subfloat[]{\includegraphics[width=0.12\textwidth]{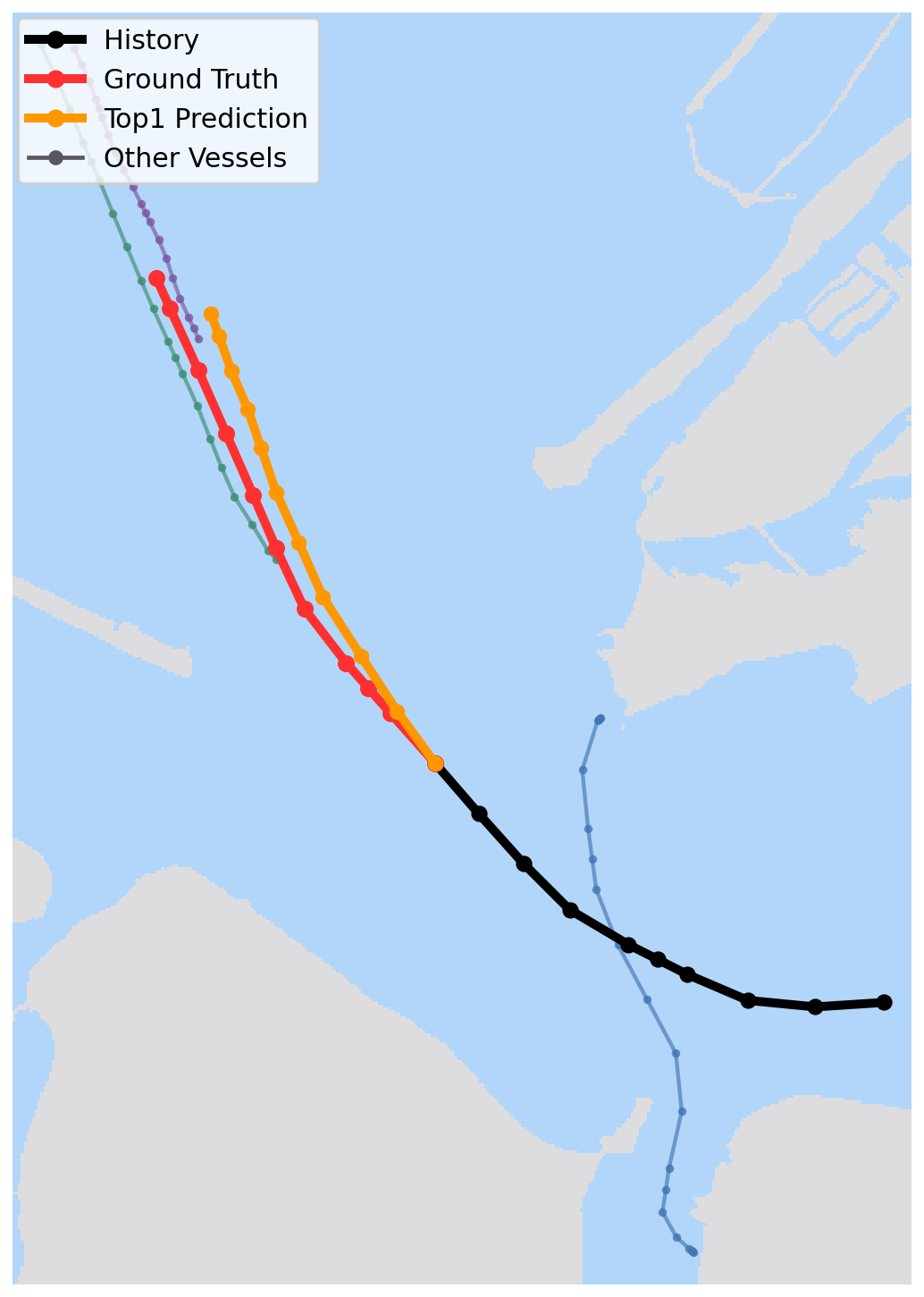}\label{fig:qubiekeshihua41}}
    \hfil
    \subfloat[]{\includegraphics[width=0.12\textwidth]{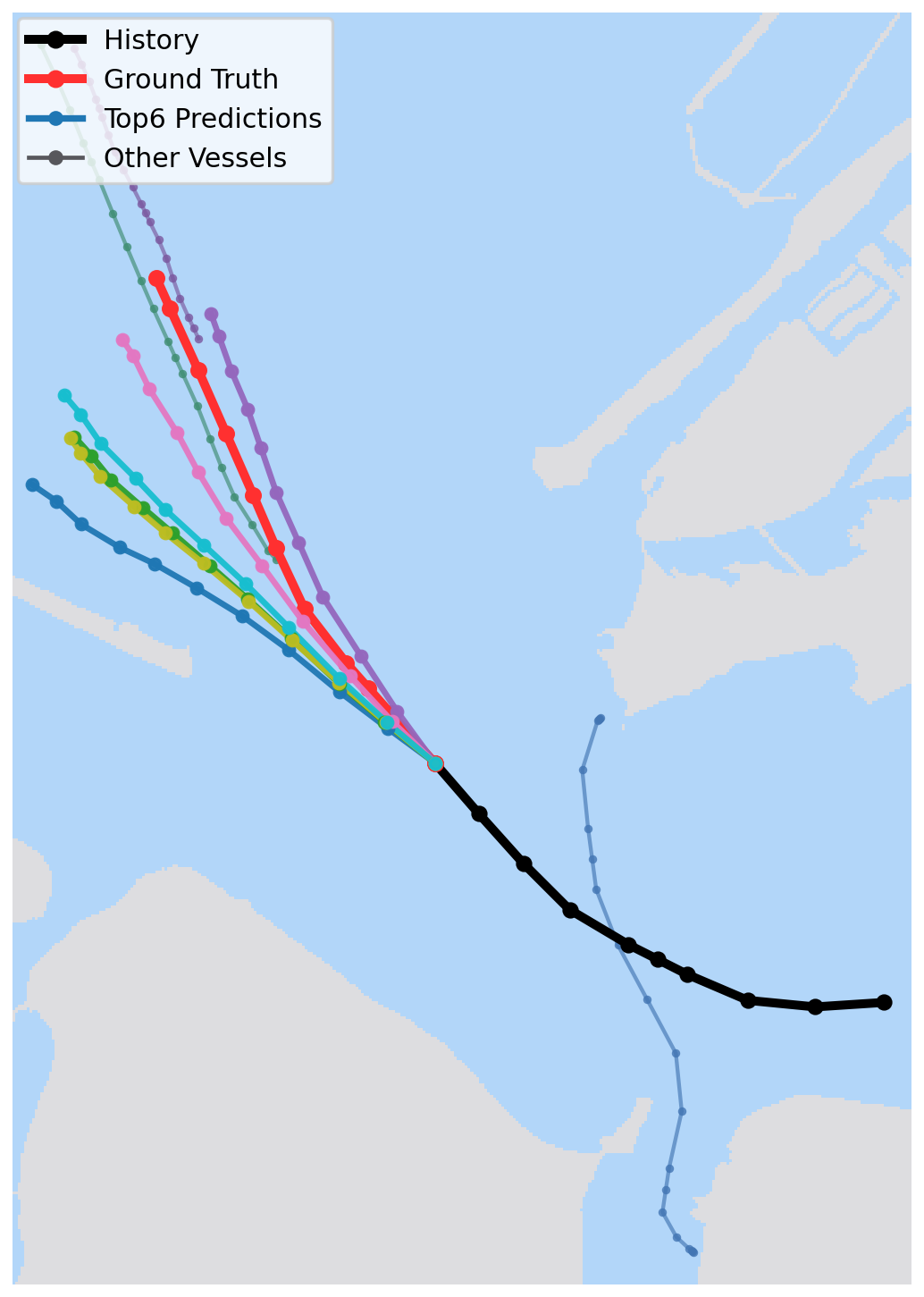}\label{fig:qubiekeshihua42}}
    \hfil
    \subfloat[]{\includegraphics[width=0.12\textwidth]{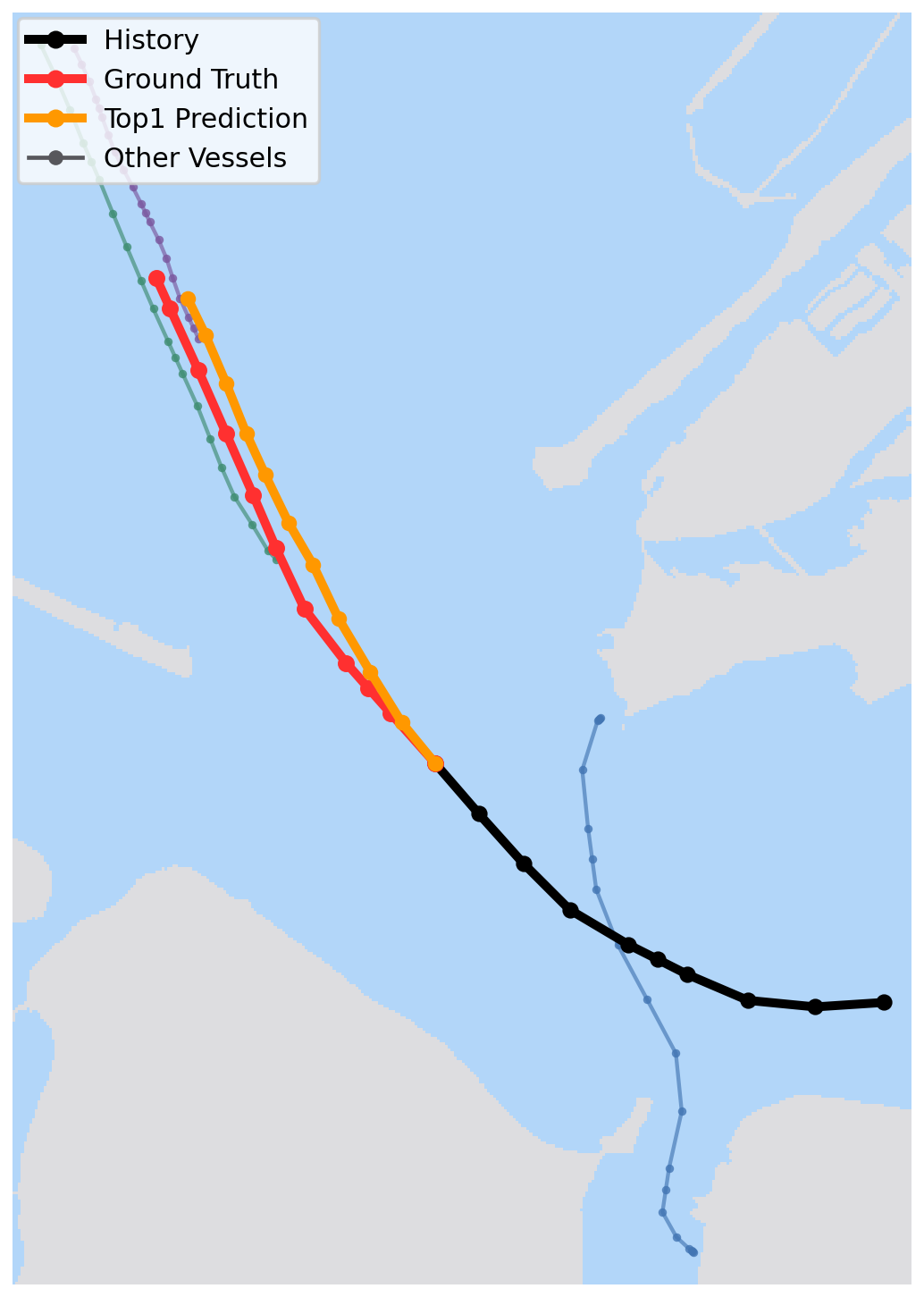}\label{fig:qubiekeshihua43}}
    \hfil
    \subfloat[]{\includegraphics[width=0.12\textwidth]{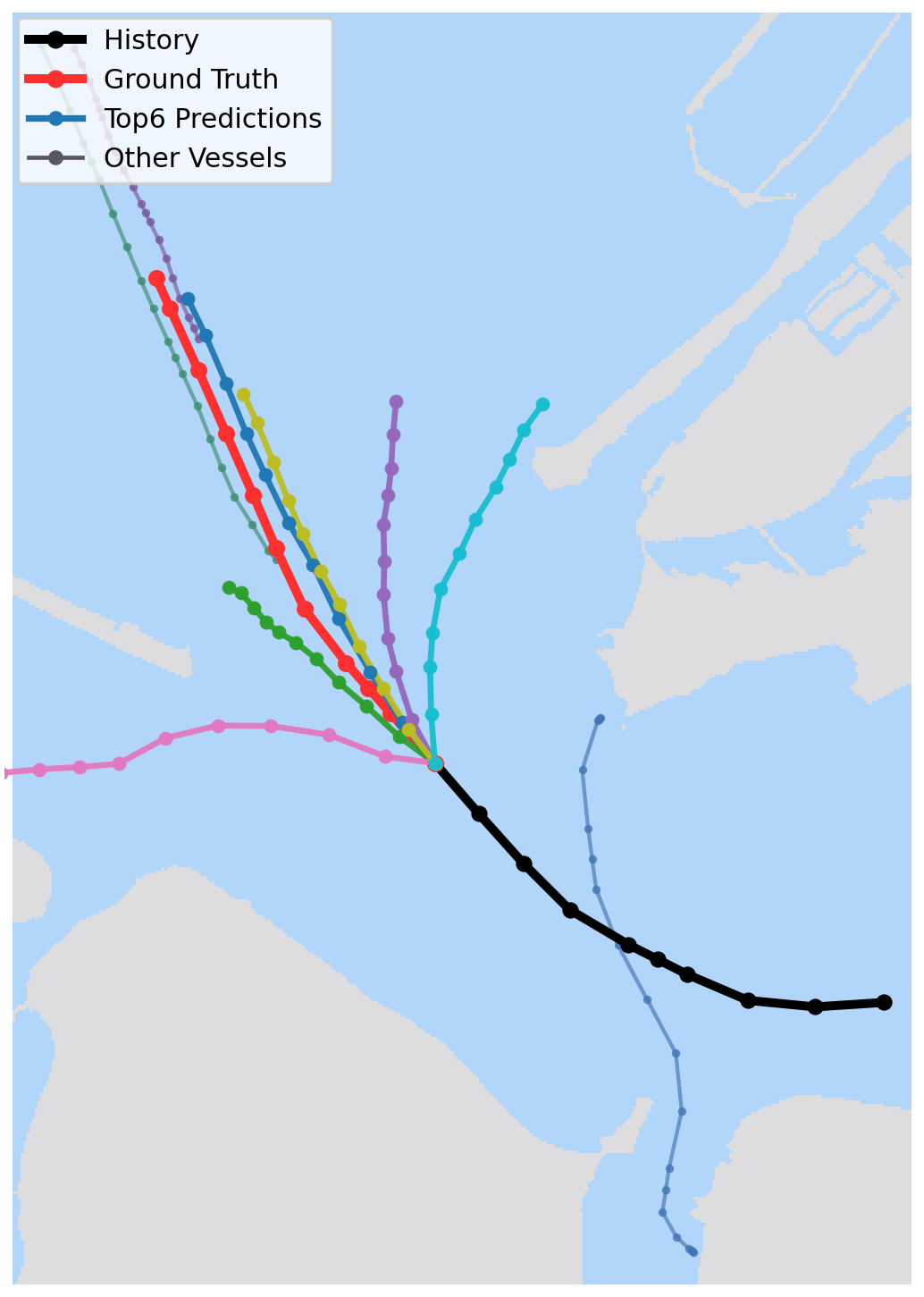}\label{fig:qubiekeshihua44}}
    \hfil
    \subfloat[]{\includegraphics[width=0.12\textwidth]{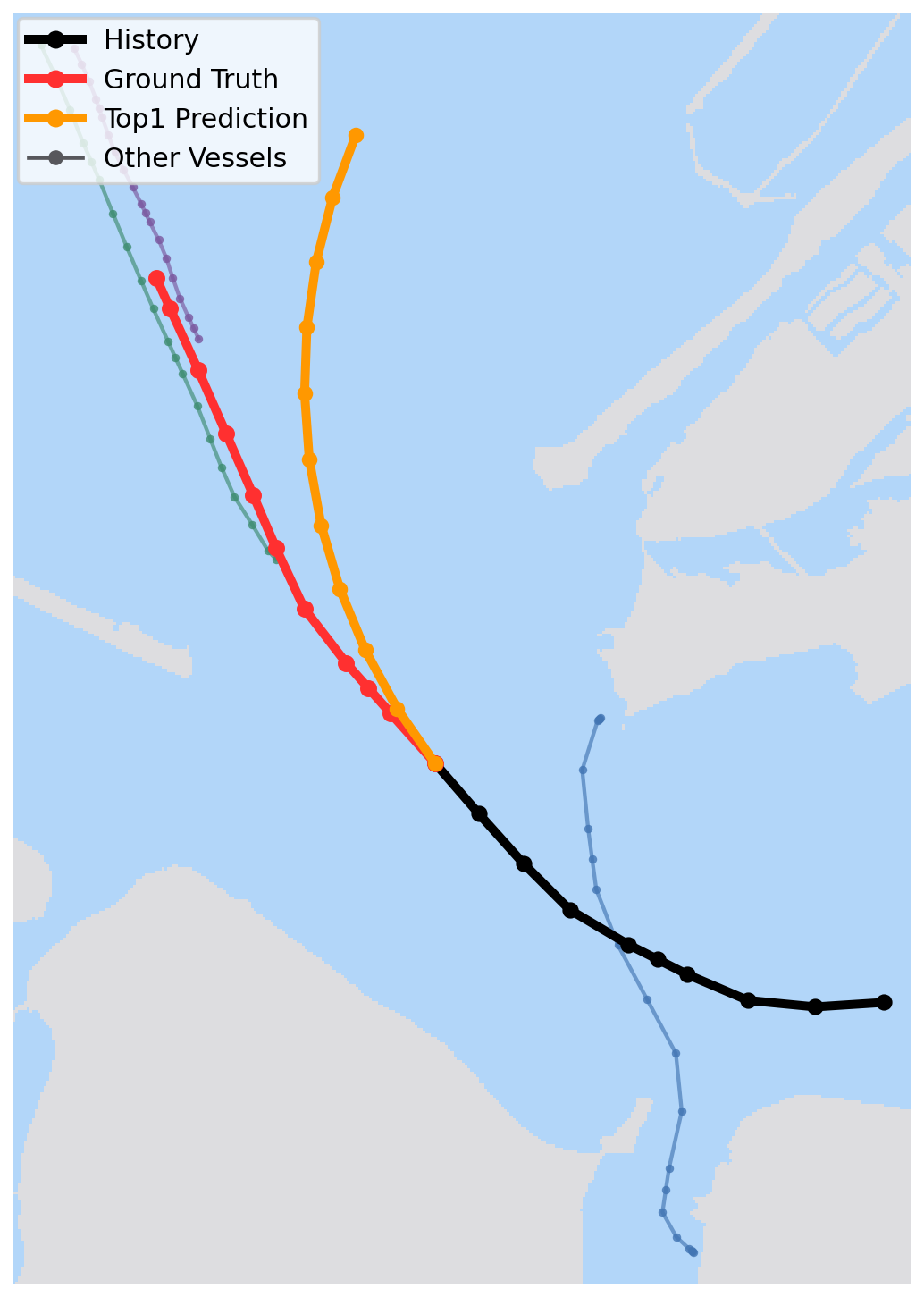}\label{fig:qubiekeshihua45}}
    \hfil
    \subfloat[]{\includegraphics[width=0.12\textwidth]{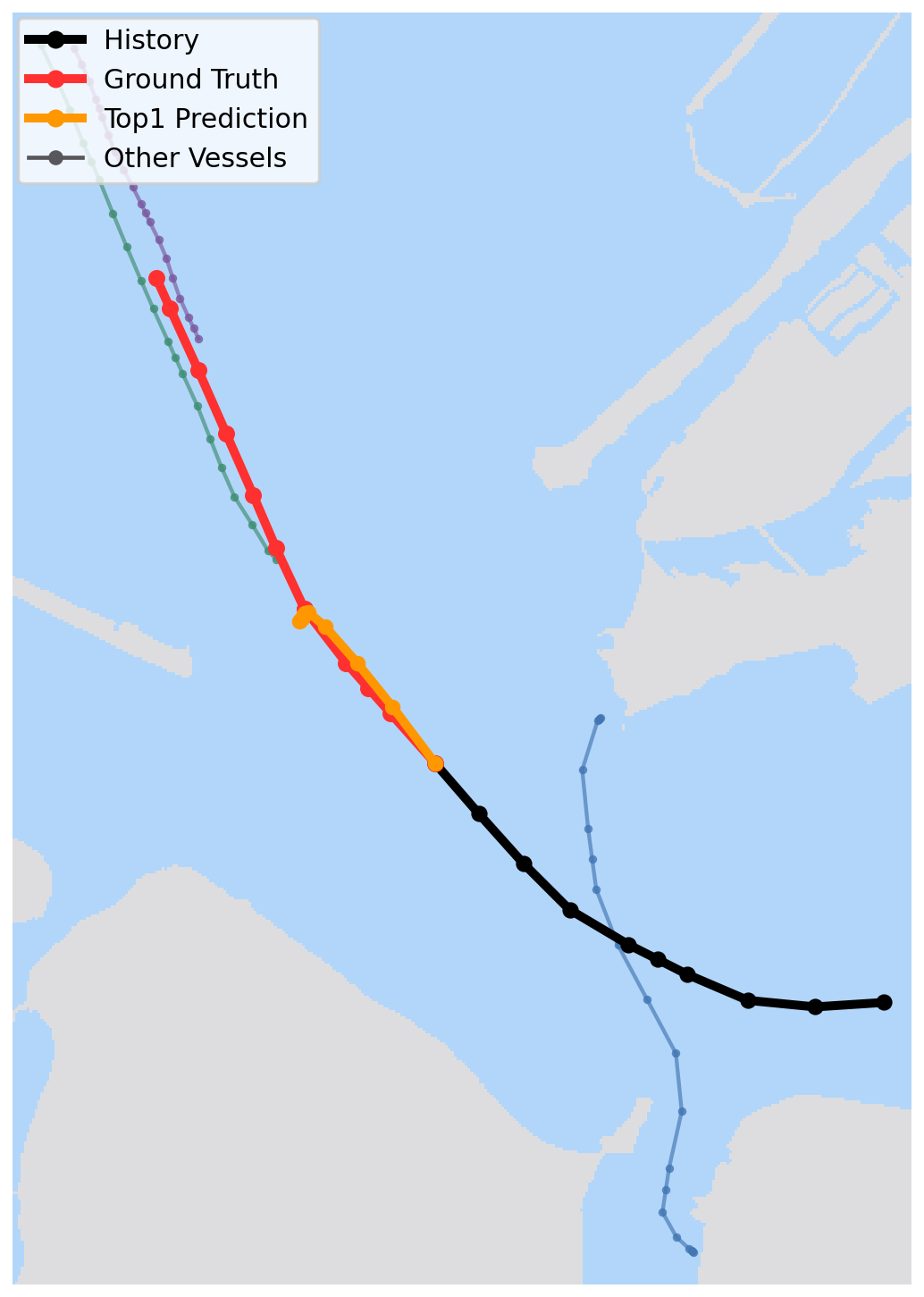}\label{fig:qubiekeshihua46}}
    \hfil
    \subfloat[]{\includegraphics[width=0.12\textwidth]{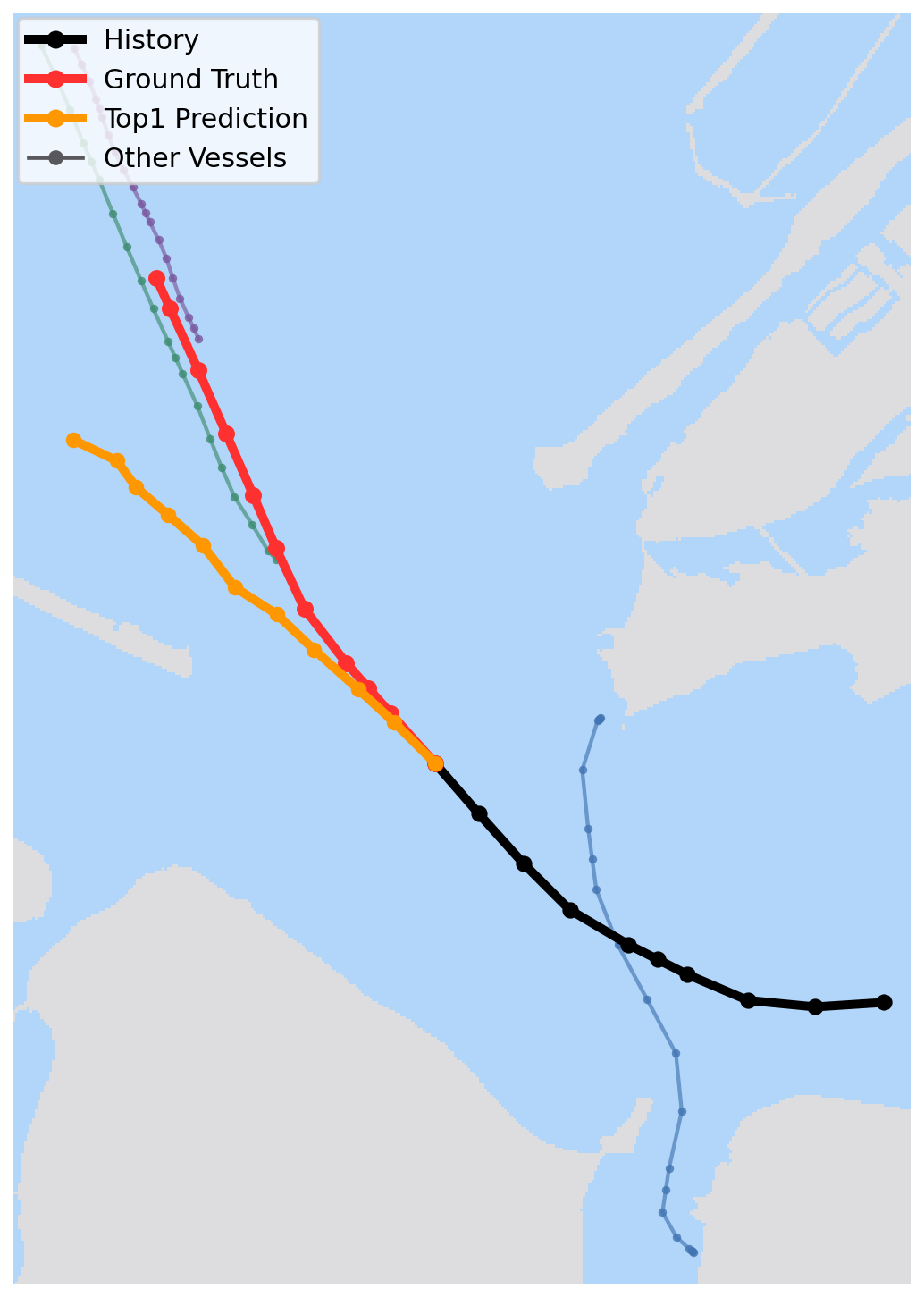}\label{fig:qubiekeshihua47}}
    \hfil
    \subfloat[]{\includegraphics[width=0.12\textwidth]{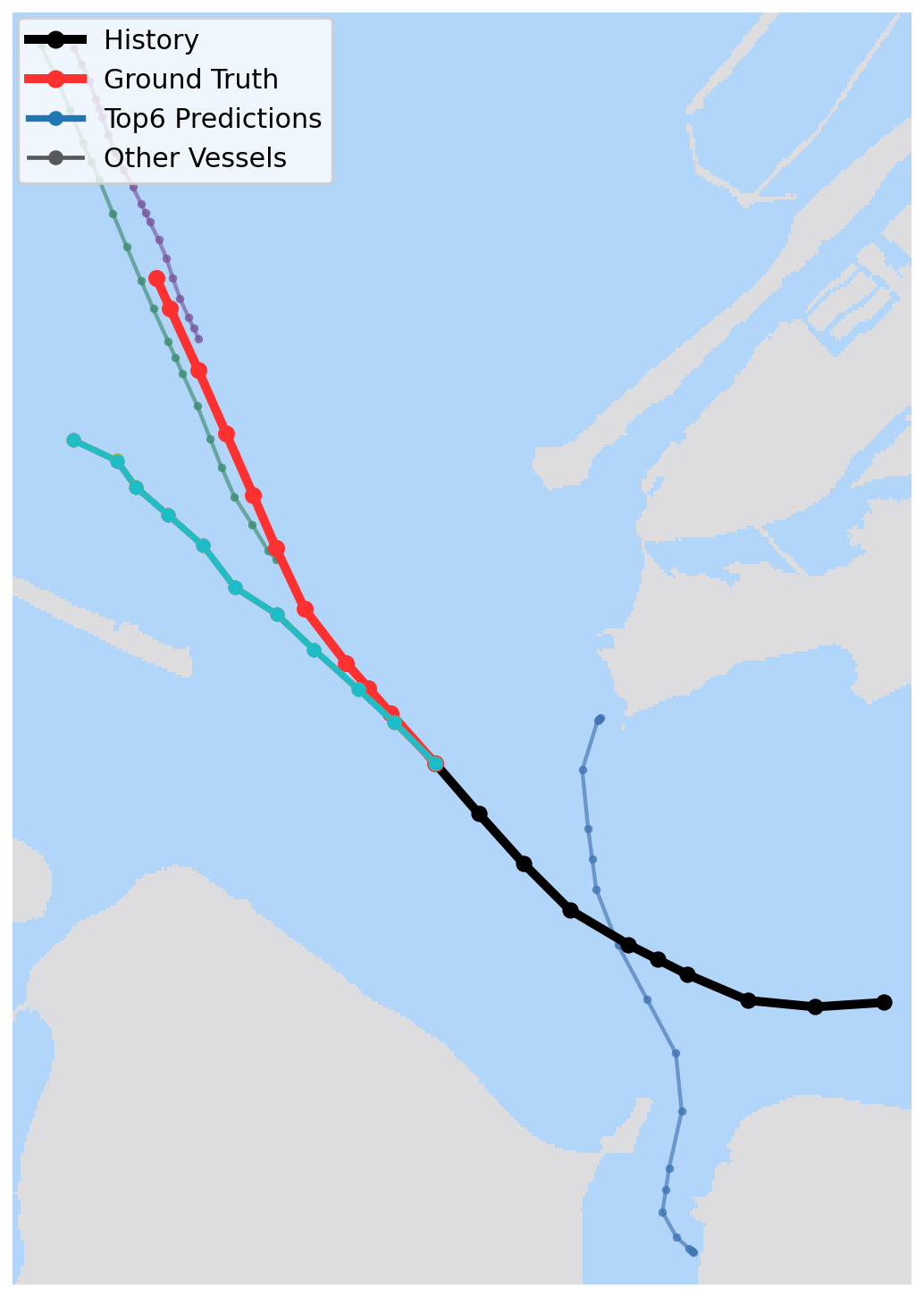}\label{fig:qubiekeshihua48}}
    \caption{Visualization of trajectory prediction results of different methods in ports and coastal areas. (a), (i): NaviLane (top-1); (b), (j): NaviLane (top-6); (c), (k): DI-MTP (top-1); (d), (l): DI-MTP (top-6); (e), (m): CTRV; (f), (n): LSTM; (g), (o): PECNet (top-1); (h), (p): PECNet (top-6).}
    \label{fig:qubiekeshihua3}
\end{figure*}

The qualitative comparisons in Fig.~\ref{fig:qubiekeshihua2} and Fig.~\ref{fig:qubiekeshihua3} provide additional insight into the behavior of different forecasting methods in inland waterways, ports, and coastal areas. Overall, NaviLane aligns more closely with the ground-truth trajectory under the top-1 setting, while under the top-6 setting it covers multiple plausible futures without producing obviously invalid candidates that deviate from the navigable channel structure. This indicates that vectorized lane priors, hierarchical macro-action modeling, and consequence-aware reranking jointly improve both the rationality and the selectability of the predicted trajectories.

A closer examination of DI-MTP reveals that its multimodal predictions are overly dispersed. Although its candidates cover a large spatial range, many of them are scattered across broad regions and fail to concentrate around high-quality futures. Moreover, several predicted trajectories are visually inconsistent with the local waterway geometry and are therefore implausible from a navigational perspective. This suggests that DI-MTP suffers from a mismatch between diversity and precision: it can spread predictions across many regions, but lacks an effective mechanism to organize them into a practically useful candidate set.

PECNet exhibits the opposite issue. Its multimodal predictions cover only a limited area, and the visual difference between the top-6 output and the single-trajectory prediction is often small. In other words, its multimodal output tends to collapse to a narrow local region, behaving more like slight perturbations of a single prediction than a set of distinct navigation strategies. This weak coverage limits its ability to capture genuinely different futures when the true trajectory deviates from the dominant mode.

By contrast, NaviLane produces multimodal candidates that are both diverse and structurally constrained. The predicted trajectories better follow curved channel geometry, remain consistent with the navigable region, and show clearer strategy-level differentiation across modes. This suggests that NaviLane does not simply increase diversity, but instead generates a more meaningful candidate set through the joint effect of lane-aware encoding, macro-action organization, local refinement, and world-model-based consequence evaluation.

Another important observation is that the advantage of NaviLane is consistent across different scenario types. In narrow inland channels, it more faithfully follows the corridor-like geometry of the waterway. In ports and coastal areas, it maintains reasonable candidate spread while still respecting local navigational structure. This consistency indicates that the model is not overfitted to a single environment type, but learns a more general forecasting mechanism across different maritime settings.

\subsection{Ablation Study}

Table~\ref{tab:ablation} presents the ablation results of different modules in the proposed framework. Removing the refinement module leads to the most severe degradation across ADE@10, FDE@10, RMSE, MaxDE, and MHE, confirming its importance for local trajectory correction. Although the macro-action decoder provides reasonable coarse futures, errors in curvature, endpoint location, and motion smoothness still remain. Without refinement, long-horizon prediction quality deteriorates sharply.

Removing the world-model-based counterfactual evaluator also causes a clear performance drop, indicating that multimodal candidate generation alone is insufficient. The world model helps distinguish scene-consistent candidates by evaluating their anticipated interaction consequences, thereby improving consequence-aware ranking.

A similar but slightly smaller degradation is observed when the selector is removed. This suggests that the selector improves the alignment between scene representations and macro-action prototypes, reducing mismatch between the current context and decoded trajectory modes. Without it, the generated futures become less consistent with the actual scene.

Overall, the three modules play complementary roles: the selector improves strategy-level mode matching, the refiner enhances local trajectory accuracy, and the world-model-based evaluator improves consequence-aware candidate ranking. The full model achieves the best performance when all components are jointly enabled.

\subsection{Hyperparameter Sensitivity Analysis}
Table~\ref{tab:hyperparameter} reports the sensitivity analysis of several key hyperparameters, including the macro-action codebook size, the number of attention heads, and the number of network layers. Overall, NaviLane remains reasonably stable across different settings, while the selected default configuration provides the best balance between representation capacity and optimization behavior.

For the macro-action codebook size, $K_{\mathrm{macro}}=128$ yields the best overall trade-off. Reducing the codebook to 64 entries weakens multimodal performance, suggesting insufficient strategy coverage, while increasing it to 256 introduces redundancy and makes prototype matching more difficult. Therefore, $K_{\mathrm{macro}}=128$ provides a suitable balance between strategy diversity and learnability.

For multi-head attention, using 8 heads produces the best results. With only 4 heads, the model may be limited in capturing different aspects of the scene, such as vessel interaction, lane geometry, and local environmental context. Increasing the number of heads to 16 does not bring further gains, indicating that simply increasing architectural complexity does not necessarily improve structured scene encoding.

A similar trend is observed in the depth analysis. Using 6 layers achieves the best overall performance, whereas 4 layers are slightly insufficient and 8 layers introduce redundancy or optimization difficulty. Overall, the hyperparameter study shows that the adopted default configuration is empirically well justified.
\begin{table}[t]
    \centering
    \caption{Ablation study of different modules in the proposed model.}
    \label{tab:ablation}
    \resizebox{\linewidth}{!}{
    \begin{tabular}{c c c c c c}
    \toprule
    Setting & ADE@10$\downarrow$ & FDE@10$\downarrow$ & RMSE$\downarrow$ & MaxDE$\downarrow$ & MHE$\downarrow$ \\
    \midrule
    w/o Module Refine & 8.07 & 15.75 & 9.50 & 15.84 & 1.35 \\
    w/o Module World-CFR & 4.10 & 8.22 & 4.84 & 8.48 & 0.48 \\
    w/o Module Selector & 3.97 & 7.81 & 4.66 & 7.99 & 0.48 \\
    All & 3.57 & 7.03 & 4.23 & 7.27 & 0.44 \\
    \bottomrule
    \end{tabular}
    }
\end{table}
\begin{table}[t]
    \centering
    \caption{Hyperparameter sensitivity analysis.}
    \label{tab:hyperparameter}
    \resizebox{\linewidth}{!}{
    \begin{tabular}{c c c c c c c}
    \toprule
    Hyperparameter & Value & ADE@10$\downarrow$ & FDE@10$\downarrow$ & minADE@10$\downarrow$ & minFDE@10$\downarrow$ & RMSE$\downarrow$ \\
    \midrule
    \multirow{3}{*}{Macro-k} 
    & $64$ & 3.55 & 6.89 & 2.75 & 5.11 & 4.18 \\
    & $128$ & 3.57 & 7.03 & 2.61 & 4.85 & 4.23 \\
    & $256$ & 3.72 & 7.21 & 2.88 & 5.39 & 4.39\\
    \midrule
    \multirow{3}{*}{num-heads} 
    & $4$ & 3.81 & 7.44 & 2.80 & 5.42 & 4.49 \\
    & $8$ & 3.57 & 7.03 & 2.61 & 4.85 & 4.23 \\
    & $16$ & 3.72 & 7.27 & 2.82 & 5.53 & 4.38 \\
    \midrule
    \multirow{3}{*}{num-layers} 
    & $4$ & 3.64 & 7.18 & 2.72 & 5.24 & 4.29 \\
    & $6$ & 3.57 & 7.03 & 2.61 & 4.85 & 4.23 \\
    & $8$ & 3.78 & 7.34 & 2.66 & 5.11 & 4.44 \\
    \bottomrule
    \end{tabular}
    }
\end{table}

\section{Conclusion}
This paper presented NaviAIS, a standardized scenario-level AIS dataset with vectorized lane priors, and NaviLane, a hierarchical macro-action forecasting framework. NaviAIS provides unified trajectory and map representations, while NaviLane integrates trajectory--map encoding, macro-action generation, refinement, and consequence-aware ranking. Experiments demonstrate its effectiveness in both single-modal and multimodal vessel trajectory prediction.
\section*{Code and Data Availability}

The source code of \textit{NaviLane} and related configuration files are publicly available for reproducibility at GitHub: 
\url{https://github.com/guiyuanal/NaviLane}. 
The \textit{NaviAIS} dataset is publicly available at Hugging Face: 
\url{https://huggingface.co/datasets/Guiyuanal/NaviAIS}.
\clearpage
\newpage
\bibliographystyle{IEEEtran}
\bibliography{references}
\end{document}